\documentclass{article}

\PassOptionsToPackage{numbers, compress}{natbib}

\usepackage[main, final]{neurips_2026}

\usepackage[utf8]{inputenc} 
\usepackage[T1]{fontenc}    
\usepackage{hyperref}       
\usepackage{url}            
\usepackage{booktabs}       
\usepackage{amsfonts}       
\usepackage{amsmath}        
\usepackage{amssymb}        
\usepackage{nicefrac}       
\usepackage{microtype}      
\usepackage{xcolor}         
\usepackage{graphicx}       
\usepackage{colortbl}       
\usepackage{caption}        
\usepackage{wrapfig}        
\usepackage{circuitikz}     
\usetikzlibrary{arrows.meta}

\title{You Don't Need Aligned Representations: Knowledge Distillation via Random Prototype Spaces}

\author{%
  Hamza Etcibasi\,$^{1,2}$ \qquad Ramazan Gokberk Cinbis\,$^{1,3}$ \\[2pt]
  $^{1}$Department of Computer Engineering, Middle East Technical University, Ankara, T\"urkiye \\
  $^{2}$ASELSAN Inc., Ankara, T\"urkiye \\
  $^{3}$ROMER Robotics and AI Center, Middle East Technical University, Ankara, Türkiye \\[2pt]
  {\small\texttt{hamza.etcibasi@metu.edu.tr}\, \texttt{haetcibasi@aselsan.com}\, \texttt{gcinbis@ceng.metu.edu.tr}}
}

\begin{document}

\maketitle

\begin{abstract}
A widely held assumption in knowledge distillation is that teacher and student representations must occupy a compatible or explicitly aligned feature space. We propose RPKD (\textbf{R}andom \textbf{P}rototype \textbf{K}nowledge \textbf{D}istillation), which discards this assumption by comparing both networks against a shared bank of randomly initialized, frozen prototype vectors. Alignment is therefore carried out in an external reference space rather than in the teacher's native feature coordinates: the teacher side is projected by fixed random maps that carry no trainable parameters, and only the student adapts.
Two branches handle the transfer: a logit branch aligning global prototype similarity distributions and a feature branch matching spatially-resolved prototype activation maps.
Both operate within the same fixed prototype space, which can be interpreted as a random feature embedding that approximately preserves inter-sample similarity structure, so the prototype bank and the distillation objective stay unchanged across teacher--student pairs while only the input projections are instantiated per architecture.
In low-category settings, decoupling the prototype vocabulary from the task label space yields richer supervision than the label space alone, an advantage absent from logit-based methods, whose supervisory signal collapses as class count shrinks. This advantage is especially relevant in real-world scenarios where the label space is typically constrained.
Extensive experiments demonstrate the effectiveness of RPKD, achieving a maximum gain of $\mathbf{+7.98\%}$ over OFA on CIFAR-100, $\mathbf{+12.44\%}$ on ImageNet-100, and $\mathbf{+1.28\%}$ on ImageNet-1K.
Source code is available at \url{https://github.com/HamzaEtcibasi/RPKD}.
\end{abstract}

\section{Introduction}

Knowledge distillation (KD)~\cite{hinton2015distilling,gou2021knowledge} transfers the predictive behavior of a large \emph{teacher} network into a compact \emph{student}, serving as a practical compression technique alongside pruning, quantization, and low-rank decomposition in standard training pipelines.
Hinton et~al.~\cite{hinton2015distilling} align student and teacher output distributions via temperature-scaled softmax logits, whose soft targets encode inter-class similarity structure (the so-called \emph{dark knowledge}) that hard labels discard.

The literature has since diverged into two directions.
\emph{Logit-based methods}~\cite{furlanello2018born,zhang2018deep,jin2019knowledge,son2021densely,huang2022knowledge,zhao2022decoupled,sun2024logit} revisit output-level objectives, showing that iterative, collaborative, or standardized soft-target losses can match feature-level transfer with far less overhead.
\emph{Feature-based methods}~\cite{romero2015fitnets,zagoruyko2016paying,yim2017gift,heo2019knowledge,heo2019comprehensive,huang2017like,peng2019correlation,park2019relational,liu2019knowledge,tian2019contrastive,ahn2019variational,lin2022knowledge,chen2022knowledge,zhang2020task} transfer intermediate representations---hints, attention maps, activation boundaries, relational structures, and contrastive embeddings---providing supervisory signal richer than soft labels alone.
The strongest results come from combining both~\cite{chen2021distilling,hao2023one}.

Nearly all these methods assume teacher and student belong to the same architectural family. This assumption is frequently violated in practice.
Modern visual recognition spans convolutional networks~\cite{he2016deep,sandler2018mobilenetv2,liu2022convnet}, vision transformers~\cite{dosovitskiy2020image,liu2021swin,touvron2021training}, and MLP-based models~\cite{tolstikhin2021mlp,touvron2022resmlp}, each with different inductive biases, feature dimensionalities, and spatial representations.
Distilling across these boundaries is not a niche scenario: an organization deploying a large pretrained ViT often has no same-family teacher for its target CNN student, making cross-architecture KD the only path short of retraining from scratch.
The obstacle is geometric. Heterogeneous feature spaces differ in dimensionality, spatial structure, and statistical properties, so standard feature-based KD is inapplicable without alignment modules that add parameters and risk introducing misalignment rather than resolving it.

Existing methods address this through \emph{learned} alignment~\cite{liu2022cross,hao2023one,zhang2025cross,li2025fuse}.
OFA~\cite{hao2023one} maps student features into the class logit space via learned stage-wise projectors; subsequent work refines this direction through redundancy suppression~\cite{zhang2025cross} or representation fusion before transfer~\cite{li2025fuse}.
Despite their effectiveness, these approaches share a structural limitation: alignment modules are architecture-specific and pair-dependent, and methods tied to the class logit space suffer supervisory signal collapse as the number of task classes shrinks.

We ask a more fundamental question: \emph{does the shared alignment space itself need to be learned, or can a fixed random space suffice?}
We propose \textbf{RPKD} (\textbf{R}andom \textbf{P}rototype \textbf{K}nowledge \textbf{D}istillation), which transfers knowledge through a shared set of randomly initialized, \emph{frozen} prototype vectors entirely decoupled from class semantics.
Both teacher and student are projected into this fixed random prototype space independently: the teacher via frozen random linear projectors that require no training, the student via a learnable MLP that adapts to match the teacher's \emph{fixed} prototype similarity patterns.
Because the shared space and all teacher-side projections are fixed at initialization, the student faces a stable, well-defined alignment target throughout training with no joint teacher--student optimization required, and the shared space has fixed dimension $M$ regardless of the number of task classes, avoiding the supervisory collapse that plagues class-logit-based methods in low-category settings.
Distillation proceeds through two complementary branches: a \emph{logit branch} that aligns global prototype similarity distributions, and a \emph{feature branch} that matches spatially-resolved prototype activation maps, jointly transferring semantic and structural knowledge within the same random space.

Crucially, the entire teacher-side construction reduces to just two lightweight frozen random linear projectors---one for global features and one for spatial features---requiring no training whatsoever; only the student's MLP adapter is optimized.
This minimal footprint makes RPKD immediately applicable to a new teacher--student pair: neither the objective nor the prototype space has to be redesigned, and the teacher side needs no training at all.
A notable practical advantage of this design emerges in low-category settings: because the prototype vocabulary is decoupled from the number of task classes, it provides consistently rich supervisory signal even when few class labels are available, a condition common in domain-specific and real-world deployment scenarios.
Our contributions are:
\begin{itemize}
  \item \textbf{RPKD}: a cross-architecture distillation framework using a frozen random prototype space, in which the shared space and all teacher-side projections are fixed at initialization and no design choice depends on the number of task classes.
  \item \textbf{Dual-path objective}: a logit branch over global prototype similarities and a feature branch over spatial activation maps, transferring both semantic and structural knowledge within a single shared space.
  \item \textbf{Low-category robustness}: an empirical and theoretical motivation for why prototype spaces provide richer, more stable supervision than the task label space when class counts are small.
  \item \textbf{State-of-the-art results}: up to $+7.98\%$ over OFA on CIFAR-100, up to $+12.44\%$ on ImageNet-100, and $+1.28\%$ on ImageNet-1K, with consistent gains across CNN and ViT architectural families and few-class subsets.
\end{itemize}

\section{Related work}
\label{sec:related}

\subsection{Vision architecture families}

\paragraph{Convolutional neural networks (CNNs).}
Convolutional neural networks~\cite{he2016deep,sandler2018mobilenetv2,liu2022convnet} have long dominated visual recognition through local connectivity and weight sharing.
Residual connections~\cite{he2016deep} enable very deep networks, depthwise-separable convolutions~\cite{sandler2018mobilenetv2} yield highly efficient variants, and ConvNeXt~\cite{liu2022convnet} modernizes the CNN design space with ideas borrowed from transformers.
CNNs remain the architecture of choice for resource-constrained deployment thanks to decades of hardware optimization.

\paragraph{Vision transformers (ViTs).}
The self-attention mechanism, originally proposed for natural language processing~\cite{vaswani2017attention}, was adapted to vision by the Vision Transformer (ViT)~\cite{dosovitskiy2020image}, which splits an image into fixed-size non-overlapping patches, linearly embeds each patch into a token, and processes the resulting sequence with a standard transformer encoder.
This formulation showed that a pure self-attention backbone matches or surpasses CNNs at scale.
DeiT~\cite{touvron2021training} addresses data efficiency through a dedicated distillation token, while Swin Transformer~\cite{liu2021swin} introduces hierarchical windowed attention that recovers spatial locality and supports dense prediction.
Beyond classification, ViTs have become the backbone of choice across detection, segmentation, generation, and vision-language tasks~\cite{carion2020end,xie2021segformer,peebles2023scalable,kirillov2023segment,radford2021learning}, yet their compute overhead motivates their use as teachers rather than deployed students.

\paragraph{Multi-layer perceptrons (MLPs).}
MLP-Mixer~\cite{tolstikhin2021mlp} and ResMLP~\cite{touvron2022resmlp} demonstrate competitive image classification using only dense matrix multiplications, replacing both convolutions and self-attention with alternating token-mixing and channel-mixing layers.
These models represent a third architectural paradigm with distinct inductive biases, further enriching the heterogeneous landscape in which cross-architecture distillation must operate.

\subsection{Knowledge distillation}

\paragraph{Logit-based distillation.}
Hinton~et~al.~\cite{hinton2015distilling} showed that temperature-scaled output distributions carry rich inter-class similarity structure as soft targets.
Born-Again Networks~\cite{furlanello2018born} demonstrate that iterative self-distillation steadily improves accuracy; Deep Mutual Learning~\cite{zhang2018deep} and densely guided distillation~\cite{son2021densely} extend this to online and multi-teacher settings.
More recently, DIST~\cite{huang2022knowledge} replaces exact distribution matching with correlation-based objectives; DKD~\cite{zhao2022decoupled} decouples the KL loss into target and non-target terms; and LSKD~\cite{sun2024logit} introduces logit standardization to remove scale sensitivity.
Because output distributions are defined over a shared class vocabulary regardless of architecture, logit-based methods are inherently architecture-agnostic; their limitation is that the supervisory signal is weaker than that of feature-level methods.
Combined approaches~\cite{chen2021distilling} that stack both objectives achieve the strongest results, but reintroduce the feature alignment challenge when teacher and student architectures differ.

\paragraph{Feature-based distillation.}
Feature-based methods transfer knowledge through intermediate representations beyond the final output.
FitNet~\cite{romero2015fitnets} introduced hint-based learning by matching compressed teacher activations; subsequent work diversified the transferred signal to attention maps~\cite{zagoruyko2016paying}, flow-of-solution matrices~\cite{yim2017gift}, activation boundaries~\cite{heo2019knowledge}, pairwise relational structures~\cite{park2019relational,liu2019knowledge,peng2019correlation}, neuron selectivity patterns~\cite{huang2017like}, contrastive embeddings~\cite{tian2019contrastive}, variational mutual information~\cite{ahn2019variational}, and transformer-aware feature transfer~\cite{lin2022knowledge}.
OFD~\cite{heo2019comprehensive} provides a principled analysis of feature distillation objectives, while SimKD~\cite{chen2022knowledge} bypasses alignment by reusing the teacher's classifier directly.
Applying these methods across heterogeneous architectures, however, demands careful design: differing feature dimensionalities, spatial structures, and inductive biases make direct intermediate feature matching ill-posed, and naive adaptation risks introducing misalignment rather than resolving it.

\paragraph{Cross-architecture distillation.}
Early works explored cross-architecture transfer in limited settings, such as CNN-to-ViT distillation~\cite{touvron2021training,chen2022dearkd} and data-efficient locality guidance~\cite{li2022locality}, but lacked a unified formulation.
More recently, OFA~\cite{hao2023one} established a general paradigm by introducing learned projectors that map heterogeneous features into a shared logit space with stage-wise supervision, enabling systematic evaluation across CNNs, ViTs, and MLPs.
Subsequent work further improves this direction by simplifying alignment through redundancy suppression~\cite{zhang2025cross} or explicitly fusing heterogeneous representations before transfer~\cite{li2025fuse}.
These methods are architecture-agnostic in the sense that they apply to heterogeneous pairs, but in all of them the shared space is \emph{learned} and anchored to one of the two networks.
RSD~\cite{zhang2025cross}, for instance, maps the student into the teacher's embedding space and suppresses redundancy within a $D_t \times D_t$ correlation matrix, so both the adapter output and the comparison space change whenever the teacher does ($D_t\!=\!192$ for DeiT-T versus $768$ for Swin-T); OFA instead anchors to the class logit space, whose width is set by the task.
RPKD differs in where the shared space comes from rather than in what it spans.
Both networks are projected into a prototype space that is random, frozen at initialization, and sized independently of the task label space, so the alignment target is fully determined before training begins and is reused unchanged across arbitrary model pairs, with only the student's input projection learned.

\section{Method}
\label{sec:method}

We propose \textbf{RPKD}, a knowledge distillation framework that transfers information through a shared \emph{random prototype space} consisting of randomly initialized, class-agnostic reference directions fixed in a common embedding space, eliminating the need for \emph{teacher-space-defined} feature alignment between heterogeneous architectures: neither network's native coordinate system serves as the target that the other must be mapped into.
Both teacher and student are projected into this space independently, and distillation is performed by matching their similarity structures within it.

\begin{figure}[!ht]
\centering
\resizebox{1\textwidth}{!}{%
\begin{circuitikz}
\tikzstyle{every node}=[font=\fontsize{18.2pt}{23.7pt}\selectfont]
\draw [ fill={rgb,255:red,235; green,235; blue,255}, fill opacity=1, line width=0.6pt ] (3.25,14.625) rectangle (11.875,13.875);
\node[anchor=center, align=center, inner xsep=0.080cm, inner ysep=0.085cm, rounded corners=0.020cm] at (7.5625,14.25) {\fontsize{18.2pt}{23.7pt}\selectfont Teacher};
\draw [ fill={rgb,255:red,255; green,235; blue,235}, fill opacity=1, line width=0.6pt ] (3.25,8) rectangle (11.875,7.25);
\node[anchor=center, align=center, inner xsep=0.080cm, inner ysep=0.085cm, rounded corners=0.020cm] at (7.5625,7.625) {\fontsize{18.2pt}{23.7pt}\selectfont Student};
\draw [ fill={rgb,255:red,255; green,235; blue,235}, fill opacity=1, line width=0.8pt ] (12.5,8.625) rectangle (13,8.125);
\draw [ fill={rgb,255:red,255; green,235; blue,235}, fill opacity=1, line width=0.8pt ] (12.5,8.125) rectangle (13,7.625);
\draw [ fill={rgb,255:red,255; green,235; blue,235}, fill opacity=1, line width=0.8pt ] (12.5,7.625) rectangle (13,7.125);
\draw [ fill={rgb,255:red,255; green,235; blue,235}, fill opacity=1, line width=0.8pt ] (12.5,7.125) rectangle (13,6.625);
\draw [ fill={rgb,255:red,235; green,235; blue,255}, fill opacity=1, line width=0.8pt ] (12.5,15.25) rectangle (13,14.75);
\draw [ fill={rgb,255:red,235; green,235; blue,255}, fill opacity=1, line width=0.8pt ] (12.5,14.75) rectangle (13,14.25);
\draw [ fill={rgb,255:red,235; green,235; blue,255}, fill opacity=1, line width=0.8pt ] (12.5,14.25) rectangle (13,13.75);
\draw [ fill={rgb,255:red,235; green,235; blue,255}, fill opacity=1, line width=0.8pt ] (12.5,13.75) rectangle (13,13.25);
\draw [ fill={rgb,255:red,231; green,230; blue,230}, fill opacity=1] (4.25,10.25) rectangle (5,9);
\node [font=\fontsize{18.2pt}{23.7pt}\selectfont, rotate around={90:(0,0)}, inner xsep=0.080cm, inner ysep=0.085cm, rounded corners=0.020cm] at (4.625,9.625) {Proj};
\draw [ fill={rgb,255:red,231; green,230; blue,230}, fill opacity=1] (5.75,10.25) rectangle (6.5,9);
\node [font=\fontsize{18.2pt}{23.7pt}\selectfont, rotate around={90:(0,0)}, inner xsep=0.080cm, inner ysep=0.085cm, rounded corners=0.020cm] at (6.125,9.625) {Proj};
\draw [ fill={rgb,255:red,231; green,230; blue,230}, fill opacity=1] (7.25,10.25) rectangle (8,9);
\node [font=\fontsize{18.2pt}{23.7pt}\selectfont, rotate around={90:(0,0)}, inner xsep=0.080cm, inner ysep=0.085cm, rounded corners=0.020cm] at (7.625,9.625) {Proj};
\draw [ fill={rgb,255:red,231; green,230; blue,230}, fill opacity=1] (8.875,10.25) rectangle (9.625,9);
\node [font=\fontsize{18.2pt}{23.7pt}\selectfont, rotate around={90:(0,0)}, inner xsep=0.080cm, inner ysep=0.085cm, rounded corners=0.020cm] at (9.25,9.625) {Proj};
\draw [-{Stealth[scale=0.5]}, ] (4.625,8) -- (4.625,9);
\draw [-{Stealth[scale=0.5]}, ] (6.125,8) -- (6.125,9);
\draw [-{Stealth[scale=0.5]}, ] (7.625,8) -- (7.625,9);
\draw [-{Stealth[scale=0.5]}, ] (9.25,8) -- (9.25,9);
\draw [-{Stealth[scale=0.5]}, ] (11.875,7.625) -- (12.5,7.625);
\draw [ fill={rgb,255:red,89; green,89; blue,89}, fill opacity=1, rounded corners = 6.0] (10.25,11.5) rectangle (11.625,10.75);
\node[anchor=center, align=center, inner xsep=0.080cm, inner ysep=0.085cm, rounded corners=0.020cm, text opacity=1] at (10.9375,11.125) {\fontsize{18.2pt}{23.7pt}\selectfont {\color[RGB]{255,255,255}$\mathcal{L}_{\mathrm{OFA}}$}};
\draw [ fill={rgb,255:red,89; green,89; blue,89}, fill opacity=1, rounded corners = 6.0] (12,10.5) rectangle (13.375,9.75);
\node[anchor=center, align=center, inner xsep=0.080cm, inner ysep=0.085cm, rounded corners=0.020cm, text opacity=1] at (12.6875,10.125) {\fontsize{18.2pt}{23.7pt}\selectfont {\color[RGB]{255,255,255}$\mathcal{L}_{\mathrm{KD}}$}};
\draw [-{Stealth[scale=0.5]}, ] (11.875,14.25) -- (12.5,14.25);
\draw [-{Stealth[scale=0.6]}, dashed] (4.625,11.125) -- (10.25,11.125);
\draw [dashed] (4.625,10.25) -- (4.625,11.125);
\draw (4.625,10.3126) -- (4.625,10.2726);
\draw (4.625,10.375) -- (4.625,10.335);
\draw (4.625,10.4376) -- (4.625,10.3976);
\draw (4.625,10.5) -- (4.625,10.46);
\draw (4.625,10.5625) -- (4.625,10.5225);
\draw (4.625,10.6875) -- (4.625,10.6475);
\draw (4.625,10.75) -- (4.625,10.71);
\draw (4.625,10.8124) -- (4.625,10.7724);
\draw (4.625,10.8749) -- (4.625,10.8349);
\draw (4.625,10.9374) -- (4.625,10.8974);
\draw (4.625,10.9999) -- (4.625,10.9599);
\draw (4.625,11.0624) -- (4.625,11.0224);
\draw [dashed] (6.125,10.25) -- (6.125,11.125);
\draw (6.125,10.3126) -- (6.125,10.2726);
\draw (6.125,10.375) -- (6.125,10.335);
\draw (6.125,10.4376) -- (6.125,10.3976);
\draw (6.125,10.5) -- (6.125,10.46);
\draw (6.125,10.5625) -- (6.125,10.5225);
\draw (6.125,10.6875) -- (6.125,10.6475);
\draw (6.125,10.75) -- (6.125,10.71);
\draw (6.125,10.8124) -- (6.125,10.7724);
\draw (6.125,10.8749) -- (6.125,10.8349);
\draw (6.125,10.9374) -- (6.125,10.8974);
\draw (6.125,10.9999) -- (6.125,10.9599);
\draw (6.125,11.0624) -- (6.125,11.0224);
\draw [dashed] (7.625,10.25) -- (7.625,11.125);
\draw (7.625,10.3126) -- (7.625,10.2726);
\draw (7.625,10.375) -- (7.625,10.335);
\draw (7.625,10.4376) -- (7.625,10.3976);
\draw (7.625,10.5) -- (7.625,10.46);
\draw (7.625,10.5625) -- (7.625,10.5225);
\draw (7.625,10.6875) -- (7.625,10.6475);
\draw (7.625,10.75) -- (7.625,10.71);
\draw (7.625,10.8124) -- (7.625,10.7724);
\draw (7.625,10.8749) -- (7.625,10.8349);
\draw (7.625,10.9374) -- (7.625,10.8974);
\draw (7.625,10.9999) -- (7.625,10.9599);
\draw (7.625,11.0624) -- (7.625,11.0224);
\draw [dashed] (9.25,10.25) -- (9.25,11.125);
\draw (9.25,10.3126) -- (9.25,10.2726);
\draw (9.25,10.375) -- (9.25,10.335);
\draw (9.25,10.4376) -- (9.25,10.3976);
\draw (9.25,10.5) -- (9.25,10.46);
\draw (9.25,10.5625) -- (9.25,10.5225);
\draw (9.25,10.6875) -- (9.25,10.6475);
\draw (9.25,10.75) -- (9.25,10.71);
\draw (9.25,10.8124) -- (9.25,10.7724);
\draw (9.25,10.8749) -- (9.25,10.8349);
\draw (9.25,10.9374) -- (9.25,10.8974);
\draw (9.25,10.9999) -- (9.25,10.9599);
\draw (9.25,11.0624) -- (9.25,11.0224);
\draw [-{Stealth[scale=0.5]}, ] (2,9.875) |- (3.25,7.625);
\draw [-{Stealth[scale=0.5]}, ] (2,11.875) |- (3.25,14.25);
\draw [-{Stealth[scale=0.6]}, dashed] (12.75,13.25) -| (12.75,10.5);
\draw [-{Stealth[scale=0.6]}, dashed] (12.75,8.625) -| (12.75,9.75);
\draw [-{Stealth[scale=0.6]}, dashed] (12.75,11.125) |- (11.625,11.125);
\node [font=\fontsize{18.2pt}{23.7pt}\selectfont, fill={rgb,255:red,255; green,255; blue,255}, fill opacity=1, text opacity=1, inner xsep=0.080cm, inner ysep=0.085cm, rounded corners=0.020cm] at (7.5,6.25) {(a) OFA};
\draw [ fill={rgb,255:red,89; green,89; blue,89}, fill opacity=1, rounded corners = 6.0] (19.125,11.25) rectangle (20.5,10.5);
\node[anchor=center, align=center, inner xsep=0.080cm, inner ysep=0.085cm, rounded corners=0.020cm, text opacity=1] at (19.8125,10.875) {\fontsize{18.2pt}{23.7pt}\selectfont {\color[RGB]{255,255,255}$\mathcal{L}_{\mathrm{feat}}$}};
\draw [-{Stealth[scale=0.5]}, ] (15.125,9.875) |- (16.125,7.625);
\node [font=\fontsize{18.2pt}{23.7pt}\selectfont, fill={rgb,255:red,255; green,255; blue,255}, fill opacity=1, text opacity=1, inner xsep=0.080cm, inner ysep=0.085cm, rounded corners=0.020cm] at (20.875,6.25) {(b) RPKD (ours)};
\draw [ fill={rgb,255:red,231; green,230; blue,230}, fill opacity=1, rotate around={90:(21.625, 8.75)}] (21.25,9.375) rectangle (22,8.125);
\node [font=\fontsize{18.2pt}{23.7pt}\selectfont, inner xsep=0.080cm, inner ysep=0.085cm, rounded corners=0.020cm] at (21.625,8.75) {$\phi_s^{\mathrm{f}}$};
\draw [ fill={rgb,255:red,231; green,230; blue,230}, fill opacity=1, rotate around={90:(23.25, 8.75)}] (22.875,9.375) rectangle (23.625,8.125);
\node [font=\fontsize{18.2pt}{23.7pt}\selectfont, inner xsep=0.080cm, inner ysep=0.085cm, rounded corners=0.020cm] at (23.25,8.75) {$\phi_s^{\mathrm{g}}$};
\draw [ fill={rgb,255:red,231; green,230; blue,230}, fill opacity=1, rotate around={90:(21.625, 13.125)}] (21.25,13.75) rectangle (22,12.5);
\node [font=\fontsize{18.2pt}{23.7pt}\selectfont, inner xsep=0.080cm, inner ysep=0.085cm, rounded corners=0.020cm] at (21.625,13.125) {$\phi_t^{\mathrm{f}}$};
\draw [ fill={rgb,255:red,231; green,230; blue,230}, fill opacity=1, rotate around={90:(23.25, 13.125)}] (22.875,13.75) rectangle (23.625,12.5);
\node [font=\fontsize{18.2pt}{23.7pt}\selectfont, inner xsep=0.080cm, inner ysep=0.085cm, rounded corners=0.020cm] at (23.25,13.125) {$\phi_t^{\mathrm{g}}$};
\draw [-{Stealth[scale=0.5]}, dashed] (21,10.875) |- (20.5,10.875);
\draw (20.9,10.875) -- (20.96,10.875);
\draw (20.8,10.875) -- (20.86,10.875);
\draw (20.7,10.875) -- (20.76,10.875);
\draw (20.6,10.875) -- (20.66,10.875);
\node (tikzmaker) [shift={(1, -1)}] at (1,11.875) {\includegraphics[width=2cm]{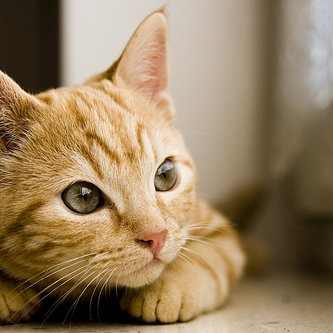}};
\draw [ fill={rgb,255:red,255; green,235; blue,235}, fill opacity=1, line width=0.6pt ] (16.125,8) rectangle (24.75,7.25);
\node[anchor=center, align=center, inner xsep=0.080cm, inner ysep=0.085cm, rounded corners=0.020cm] at (20.4375,7.625) {\fontsize{18.2pt}{23.7pt}\selectfont Student};
\draw [ fill={rgb,255:red,235; green,235; blue,255}, fill opacity=1, line width=0.6pt ] (16.125,14.625) rectangle (24.75,13.875);
\node[anchor=center, align=center, inner xsep=0.080cm, inner ysep=0.085cm, rounded corners=0.020cm] at (20.4375,14.25) {\fontsize{18.2pt}{23.7pt}\selectfont Teacher};
\draw [ fill={rgb,255:red,235; green,235; blue,255}, fill opacity=1, line width=0.8pt ] (25.375,15.25) rectangle (25.875,14.75);
\draw [ fill={rgb,255:red,235; green,235; blue,255}, fill opacity=1, line width=0.8pt ] (25.375,14.75) rectangle (25.875,14.25);
\draw [ fill={rgb,255:red,235; green,235; blue,255}, fill opacity=1, line width=0.8pt ] (25.375,14.25) rectangle (25.875,13.75);
\draw [ fill={rgb,255:red,235; green,235; blue,255}, fill opacity=1, line width=0.8pt ] (25.375,13.75) rectangle (25.875,13.25);
\draw [-{Stealth[scale=0.5]}, ] (24.75,14.25) -- (25.375,14.25);
\draw [ fill={rgb,255:red,255; green,235; blue,235}, fill opacity=1, line width=0.8pt ] (25.375,8.625) rectangle (25.875,8.125);
\draw [ fill={rgb,255:red,255; green,235; blue,235}, fill opacity=1, line width=0.8pt ] (25.375,8.125) rectangle (25.875,7.625);
\draw [ fill={rgb,255:red,255; green,235; blue,235}, fill opacity=1, line width=0.8pt ] (25.375,7.625) rectangle (25.875,7.125);
\draw [ fill={rgb,255:red,255; green,235; blue,235}, fill opacity=1, line width=0.8pt ] (25.375,7.125) rectangle (25.875,6.625);
\draw [-{Stealth[scale=0.5]}, ] (24.75,7.625) -- (25.375,7.625);
\draw [ fill={rgb,255:red,89; green,89; blue,89}, fill opacity=1, rounded corners = 6.0] (24.5,11.25) rectangle (25.875,10.5);
\node[anchor=center, align=center, inner xsep=0.080cm, inner ysep=0.085cm, rounded corners=0.020cm, text opacity=1] at (25.1875,10.875) {\fontsize{18.2pt}{23.7pt}\selectfont {\color[RGB]{255,255,255}$\mathcal{L}_{\mathrm{logit}}$}};
\draw [-{Stealth[scale=0.5]}, dashed] (24,10.875) |- (24.5,10.875);
\draw (24.1,10.875) -- (24.04,10.875);
\draw (24.2,10.875) -- (24.14,10.875);
\draw (24.3,10.875) -- (24.24,10.875);
\draw (24.4,10.875) -- (24.34,10.875);
\draw [-{Stealth[scale=0.5]}, ] (15.125,11.875) |- (16.125,14.25);
\node (tikzmaker) [shift={(1, -1)}] at (14.125,11.875) {\includegraphics[width=2cm]{tabby}};
\draw [ color={rgb,255:red,255; green,114; blue,0}, draw opacity=1 , fill={rgb,255:red,248; green,182; blue,150}, fill opacity=1, rounded corners = 6.6] (26.375,11.625) rectangle (29.375,10.125);
\node [font=\fontsize{11.9pt}{15.5pt}\selectfont, fill={rgb,255:red,248; green,182; blue,150}, fill opacity=1, text opacity=1, inner xsep=0.080cm, inner ysep=0.085cm, rounded corners=0.020cm] at (27.875,11.25) {$\mathcal{L}_{\mathrm{logit}} + \mathcal{L}_{\mathrm{feat}}$};
\node [font=\fontsize{11.9pt}{15.5pt}\selectfont, color={rgb,255:red,255; green,0; blue,0}, text opacity=1, fill={rgb,255:red,248; green,182; blue,150}, fill opacity=1, inner xsep=0.080cm, inner ysep=0.085cm, rounded corners=0.020cm] at (27.875,10.5) {\textbf{RPKD Loss}};
\node [font=\fontsize{11.9pt}{15.5pt}\selectfont, color={rgb,255:red,255; green,0; blue,0}, text opacity=1, fill={rgb,255:red,248; green,182; blue,150}, fill opacity=1, inner xsep=0.080cm, inner ysep=0.085cm, rounded corners=0.000cm] at (27.875,10.875) {\textbf{=}};
\node (tikzmaker) [shift={(1.9375, -1.9375)}] at (20.5,12.75) {\includegraphics[width=3.875cm]{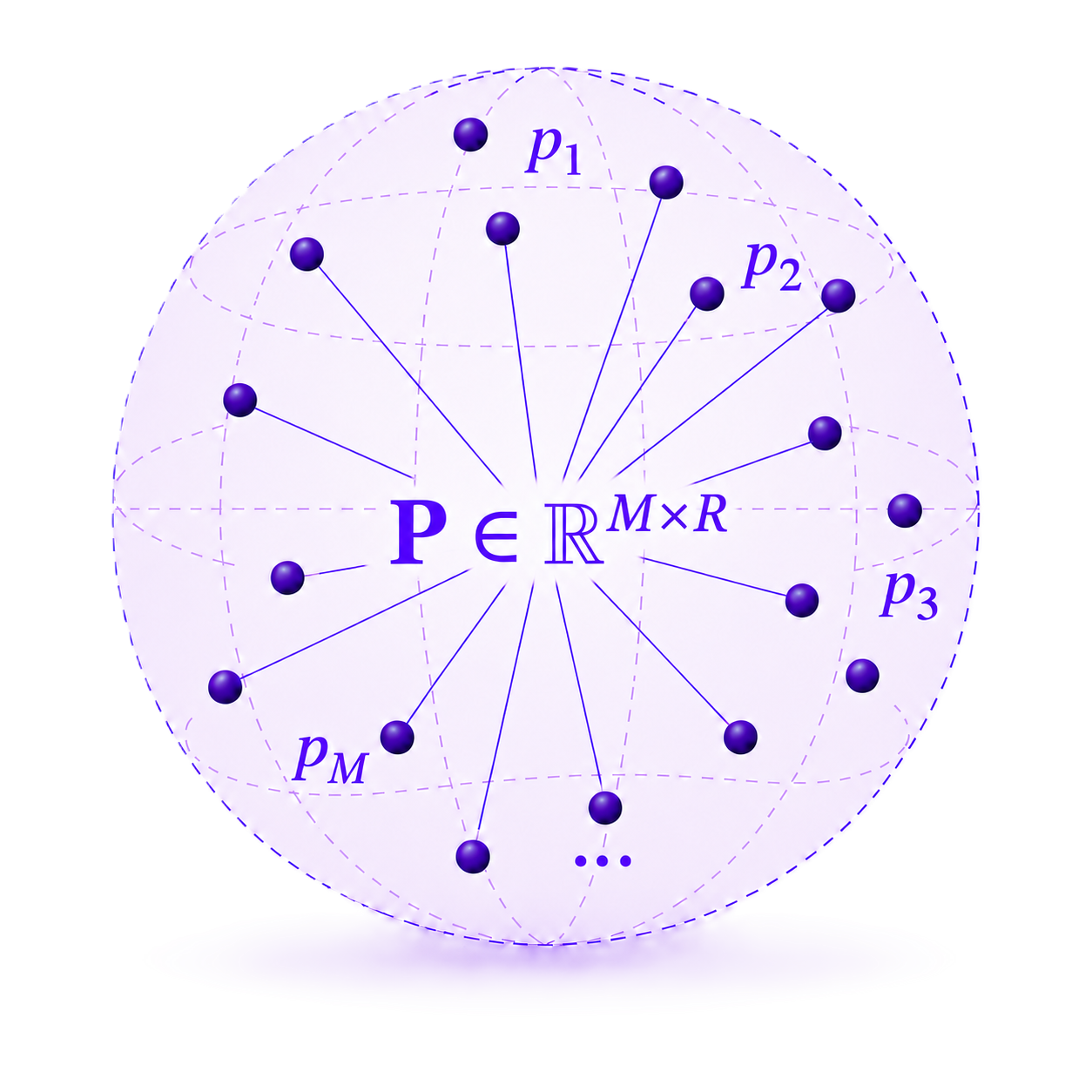}};
\node (tikzmaker) [shift={(0.3125, -0.3125)}] at (20.75,13.75) {\includegraphics[width=0.625cm]{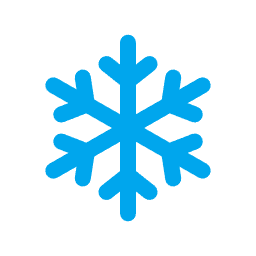}};
\node (tikzmaker) [shift={(0.3125, -0.3125)}] at (22.375,13.75) {\includegraphics[width=0.625cm]{snow}};
\node (tikzmaker) [shift={(0.25, -0.25)}] at (20.875,9.25) {\includegraphics[width=0.5cm]{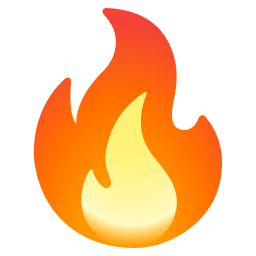}};
\node (tikzmaker) [shift={(0.25, -0.25)}] at (22.5,9.25) {\includegraphics[width=0.5cm]{fire}};

\draw [-{Stealth[scale=0.5]}] (22.4375,8) -- (22.4375,8.1875) -- (21.625,8.1875) -- (21.625,8.375);
\draw [-{Stealth[scale=0.5]}] (22.4375,8) -- (22.4375,8.1875) -- (23.25,8.1875) -- (23.25,8.375);

\draw [-{Stealth[scale=0.5]}] (22.4375,13.875) -- (22.4375,13.6875) -- (21.625,13.6875) -- (21.625,13.5);
\draw [-{Stealth[scale=0.5]}] (22.4375,13.875) -- (22.4375,13.6875) -- (23.25,13.6875) -- (23.25,13.5);

\draw [-{Stealth[scale=0.5]}] (21.625,9.125) -- (21.625,9.25) -- (22.4375,9.25) -- (22.4375,9.4);
\draw [-{Stealth[scale=0.5]}] (23.25,9.125) -- (23.25,9.25) -- (22.4375,9.25) -- (22.4375,9.4);

\draw [-{Stealth[scale=0.5]}] (21.625,12.75) -- (21.625,12.625) -- (22.4375,12.625) -- (22.4375,12.5);
\draw [-{Stealth[scale=0.5]}] (23.25,12.75) -- (23.25,12.625) -- (22.4375,12.625) -- (22.4375,12.5);

\end{circuitikz}
}%
\caption{
  \textbf{Method comparison: OFA vs.\ RPKD.}
  \textbf{(a) OFA} places four learned Proj modules at intermediate student stages, each mapping student features into the shared \emph{class logit space} ($C$-dimensional) for stage-wise alignment with the teacher's output distribution; projectors are trained jointly and are architecture-specific.
  \textbf{(b) RPKD} transfers knowledge through a \emph{frozen random} prototype matrix $\mathbf{P}\!\in\!\mathbb{R}^{M\times R}$, fixed at initialization and shared by both networks.
  The teacher projects into $\mathbf{P}$ via frozen random linear maps $\phi_t^{\mathrm{g}}, \phi_t^{\mathrm{f}}$ (\raisebox{-1pt}{\includegraphics[height=8pt]{snow}}~frozen), requiring no teacher-side updates.
  The student adapts via learnable MLP projectors $\phi_s^{\mathrm{g}}, \phi_s^{\mathrm{f}}$ (\raisebox{-1pt}{\includegraphics[height=8pt]{fire}}~learnable) to match the teacher's prototype similarity patterns.
  Two complementary branches are optimized alongside the task cross-entropy: $\mathcal{L}_{\mathrm{logit}}$ (global prototype similarity) and $\mathcal{L}_{\mathrm{feat}}$ (spatial prototype activation maps).
  Unlike OFA, whose projectors target a space whose width is set by the task, RPKD keeps the prototype bank and both losses unchanged across pairs; only the two input projections are sized to each network's feature width.
}
\label{fig:method_comparison}
\end{figure}

\paragraph{Random prototype space.}
We define a prototype matrix $\mathbf{P} \in \mathbb{R}^{M \times R}$, where $M$ is the number of prototypes and $R$ is the shared embedding dimension.
Rows of $\mathbf{P}$ are sampled independently from a standard Gaussian and L2-normalized; $\mathbf{P}$ is fixed after initialization and contains no learnable parameters, removing any dependency on class semantics or architecture-specific inductive biases.

To project representations into this space, we use separate mappings for teacher and student.
The teacher uses frozen random linear projections $\phi_t^{\mathrm{g}}, \phi_t^{\mathrm{f}} : \mathbb{R}^{D_t} \to \mathbb{R}^R$ (where $D_t$ is the teacher feature dimension), while the student uses learnable two-layer MLP projections $\phi_s^{\mathrm{g}}, \phi_s^{\mathrm{f}} : \mathbb{R}^{D_s} \to \mathbb{R}^R$ with GELU activations.
This asymmetry gives the student a fixed, stable target: the teacher's prototype similarity distribution does not shift during training.

\paragraph{Architecture-agnostic feature routing.}
We use each network's native classification feature as the global representation:
global average pooling for CNNs, the CLS token for ViT-style models~\cite{dosovitskiy2020image,touvron2021training},
and mean pooling over output tokens for models without a dedicated CLS token
(e.g., Swin Transformer~\cite{liu2021swin}, MLP-Mixer~\cite{tolstikhin2021mlp}, ResMLP~\cite{touvron2022resmlp}).
Spatial representations are the corresponding spatial feature maps or token sequences from the final network stage,
reshaped to a common $B \times N \times D$ format ($B$: batch size, $N$: number of spatial positions or tokens).
Both branches operate on \emph{final-stage} features by default; Section~\ref{sec:experiments} ablates this choice and confirms that including earlier stages degrades performance, since shallow features carry low-level spatial statistics that are difficult to reconcile across heterogeneous architectures through prototype similarity alone.

\paragraph{Distillation in prototype space.}
RPKD transfers knowledge through two complementary objectives.

\emph{Logit branch.}
Let $\mathbf{f}_g^{(x)} \in \mathbb{R}^{D_x}$ be the global feature of model $x \in \{t,s\}$.
We project it to $\mathbb{R}^R$, L2-normalize, and compute its cosine similarity to each prototype:
\begin{equation}
\mathbf{z}^{(x)} = \mathrm{norm}\!\left(\phi_x^{\mathrm{g}}(\mathbf{f}_g^{(x)})\right) \in \mathbb{R}^R, \qquad
\mathbf{l}^{(x)} = \mathbf{z}^{(x)} \mathbf{P}^\top \in \mathbb{R}^M.
\end{equation}
We refer to $\mathbf{l}^{(x)}$ as \emph{prototype logits}: despite the name, its $M$ entries are similarities to the prototypes, not scores over the $C$ task classes, and this branch never enters the class space.
That is precisely what keeps the width of the transferred signal fixed at $M$ regardless of $C$.
We match the teacher and student prototype-logit distributions with a temperature-scaled KL divergence, applying the per-sample logit standardization of LSKD~\cite{sun2024logit} to each side beforehand:
\begin{equation}
\mathcal{L}_{\mathrm{logit}} =
T^2 \cdot \mathrm{KL}\!\left(
\sigma\!\left(\tfrac{\tilde{\mathbf{l}}^{(s)}}{T}\right)
\;\Big\|\;
\sigma\!\left(\tfrac{\tilde{\mathbf{l}}^{(t)}}{T}\right)
\right),
\end{equation}
where $\tilde{\mathbf{l}}^{(x)} = (\mathbf{l}^{(x)} - \mu_x)/s_x$, with $\mu_x$ and $s_x$ the mean and standard deviation of the $M$ entries of $\mathbf{l}^{(x)}$ for that sample, $T > 0$ the distillation temperature, and $\sigma(\cdot)$ the softmax.
Standardizing each side separately removes the scale and spread mismatch between the teacher's and the student's similarity vectors, so the KL term compares the shape of the two prototype-logit distributions rather than their magnitude.

\emph{Feature branch.}
Let $\mathbf{F}^{(x)} \in \mathbb{R}^{B \times N \times D_x}$ be the spatial token matrix ($N$ spatial positions or patch tokens, each of dimension $D_x$).
We project each token to $\mathbb{R}^R$, L2-normalize, and compute per-token prototype similarities:
\begin{equation}
\mathbf{G}^{(x)} = \mathrm{norm}\!\left(\phi_x^{\mathrm{f}}(\mathbf{F}^{(x)})\right) \in \mathbb{R}^{B \times N \times R}, \qquad
\mathbf{A}^{(x)} = \mathbf{G}^{(x)} \mathbf{P}^\top \in \mathbb{R}^{B \times N \times M}.
\end{equation}
Each of the $M$ columns of $\mathbf{A}^{(x)}$ is reshaped into a spatial map and resized to $7{\times}7$, yielding prototype activation maps $\mathbf{a}^{(x)}_{b,m} \in \mathbb{R}^{49}$ for sample $b$ and prototype $m$.
Every map is then standardized over its own spatial extent, $\hat{\mathbf{a}}^{(x)}_{b,m} = (\mathbf{a}^{(x)}_{b,m} - \mu^{(x)}_{b,m})/(s^{(x)}_{b,m} + \epsilon)$ with $\mu$ and $s$ the mean and standard deviation over the $49$ positions, and teacher and student maps are aligned via a cosine similarity loss:
\begin{equation}
\mathcal{L}_{\mathrm{feat}} =
\frac{1}{BM} \sum_{b=1}^{B} \sum_{m=1}^{M}
\left(1 - \cos\!\left(\hat{\mathbf{a}}^{(t)}_{b,m},\, \hat{\mathbf{a}}^{(s)}_{b,m}\right)\right).
\end{equation}
Because the cosine is scale invariant, the division only conditions the computation numerically; it is the centring that changes the objective, making each per-map term a Pearson correlation between the teacher's and the student's activation pattern rather than a raw cosine.\footnote{For Swin students we compare the maps without this standardization, which otherwise destabilizes their training.}
\paragraph{Training objective.}
The overall loss combines cross-entropy with both distillation branches:
\begin{equation}
\mathcal{L} =
\lambda_{\mathrm{CE}}\,\mathcal{L}_{\mathrm{CE}} +
\lambda_{\mathrm{logit}}\,\mathcal{L}_{\mathrm{logit}} +
\lambda_{\mathrm{feat}}\,\mathcal{L}_{\mathrm{feat}}.
\end{equation}
Gradients are applied only to the student encoder and its projection heads.
The teacher network, prototype matrix, and teacher projectors remain fixed throughout training.

\paragraph{Discussion.}
RPKD replaces teacher-space-defined feature alignment with similarity matching in a shared random space that belongs to neither network.
Original feature spaces cannot serve as the shared medium: they differ in dimensionality ($D_t \neq D_s$), inductive bias (local vs.\ global vs.\ token-mixing), and distributional properties, making direct similarity comparison ill-posed.
A random projection space resolves this by providing an external reference frame independent of both architectures: by the Johnson--Lindenstrauss lemma~\cite{johnson1984extensions,dasgupta2003elementary}, a random linear map approximately preserves pairwise inner products between unit-norm vectors with high probability, so the similarity structure meaningful to the teacher survives the projection.
Concretely, for any two L2-normalized projected features $\mathbf{z}_1, \mathbf{z}_2 \in \mathbb{R}^R$ and prototype rows $\mathbf{p}_m$ drawn uniformly from the unit sphere,
\begin{equation}
  \mathbb{E}\!\left[(\mathbf{z}_1 \cdot \mathbf{p}_m)(\mathbf{z}_2 \cdot \mathbf{p}_m)\right] \;\propto\; \mathbf{z}_1 \cdot \mathbf{z}_2;
\end{equation}
by concentration, $\tfrac{1}{M}\,\mathbf{l}^{(t)}(x_1)\cdot\mathbf{l}^{(t)}(x_2) \approx \mathbf{z}^{(t)}(x_1)\cdot\mathbf{z}^{(t)}(x_2)$ holds simultaneously for all input pairs with high probability as $M$ grows.
As a consequence, the inter-sample similarity structure of the teacher's representations is faithfully reflected in the random prototype space: two inputs that are semantically close in the teacher's feature space produce similar prototype activation vectors.
The student, by aligning its activation distribution to the teacher's, implicitly learns to reproduce this structural geometry without direct access to the teacher's feature space, which is what decouples distillation from architectural constraints.

\section{Experiments}
\label{sec:experiments}

We evaluate RPKD on CIFAR-100 cross-architecture benchmarks spanning CNN, ViT, and MLP architectural families, on full-scale ImageNet-1K, and on ImageNet subsets designed specifically to stress the low-category regime discussed in Section~\ref{sec:method}.

\paragraph{Setup.}
CIFAR-100~\cite{krizhevsky2009learning} experiments cover 10 teacher--student pairs across two architectural families.
\emph{CNN-based students} (ResNet18~\cite{he2016deep}, MobileNetV2~\cite{sandler2018mobilenetv2}) are paired with transformer and MLP teachers (Swin-T~\cite{liu2021swin}, ViT-S~\cite{dosovitskiy2020image}, Mixer-B/16~\cite{tolstikhin2021mlp}).
\emph{ViT-based students} (DeiT-T~\cite{touvron2021training}, Swin-P\footnote{Swin-P (Swin-Pico) and Swin-N (Swin-Nano) are compact modified variants of Swin-T~\cite{liu2021swin} introduced by OFA~\cite{hao2023one}; we follow their architecture definitions for fair comparison.}) are paired with CNN and MLP teachers (ConvNeXt-T~\cite{liu2022convnet}, Mixer-B/16).
All teacher models follow the pretrained checkpoints provided by OFA~\cite{hao2023one} for fair comparison and are frozen during distillation.
All images are resized to $224{\times}224$.
For CIFAR-100, students are trained for 300 epochs using SGD with a batch size of 128, an initial learning rate of 0.05, and weight decay of $2{\times}10^{-3}$.
For ImageNet-100 and ImageNet-1K, CNN students are trained for 100 epochs using AdamW with a batch size of 64, an initial learning rate of $3{\times}10^{-3}$, weight decay of $3{\times}10^{-2}$, and 3 warmup epochs.
ViT-based students on ImageNet-100 use AdamW with a batch size of 128, a learning rate of $5{\times}10^{-4}$, weight decay of $0.05$, and 20 warmup epochs.
All settings use a cosine learning rate schedule.
RPKD uses $M\!=\!500$ randomly initialized frozen prototypes of dimension $R\!=\!256$ (ablated in Table~\ref{tab:ablation_mr}), distillation temperature $T\!=\!4.0$, and loss coefficients selected following common practice in prior KD works and validated via a small grid search.
Teacher projectors are initialized with weight standard deviation $\sigma\!=\!0.02$.
All experiments are implemented in PyTorch~\cite{paszke2019pytorch} and run on four NVIDIA H100 GPUs.

\paragraph{Baselines.}
We compare against two groups of methods.
\emph{Hint-based methods} (FitNet~\cite{romero2015fitnets}, CC~\cite{peng2019correlation}, RKD~\cite{park2019relational}, and CRD~\cite{tian2019contrastive}) transfer intermediate features or relational structures; for cross-architecture pairs they are equipped with lightweight linear adapters following the protocol of OFA~\cite{hao2023one}.
\emph{Logit-based methods} (KD~\cite{hinton2015distilling}, DKD~\cite{zhao2022decoupled}, and DIST~\cite{huang2022knowledge}) operate on output distributions and require no feature alignment.
\emph{Cross-architecture methods} are the closest competitors: OFA~\cite{hao2023one} uses a jointly optimized universal feature bridge, FBT~\cite{li2025fuse} fuses heterogeneous representations before transfer, and RSD~\cite{zhang2025cross} simplifies alignment through redundancy suppression.
For FBT and RSD we report the numbers published by their authors; benchmarks not covered by the original paper are omitted from the corresponding table.
Neither OFA nor RSD reports ImageNet-100 results, so for that benchmark we ran both ourselves from the official implementations over three seeds (RSD with a single fixed configuration, $\lambda\!=\!0.1$, $\kappa\!=\!0.01$, $\gamma\!=\!2$), and FBT is omitted because its implementation is not public.
All other baselines and training hyperparameters follow those of OFA~\cite{hao2023one} for a fair comparison.
In Tables~\ref{tab:cross_kd_cifar100}, \ref{tab:cross_kd_imagenet100}, and~\ref{tab:cross_kd_imagenet}, these methods and RPKD are grouped under \textbf{CAKD} (Cross-Architecture Knowledge Distillation), denoting methods designed to operate across heterogeneous architectural families.
RPKD results are reported as mean $\pm$ std over three independent runs on all three benchmarks; the ImageNet-100 OFA and RSD entries are likewise three-seed means, and all remaining baseline numbers are the single-run figures published by their authors.

\begin{table}[t]
\centering
\setlength{\tabcolsep}{1pt}
\small
\caption{Cross-architecture KD results on CIFAR-100. Best results in \textbf{bold}, second best \underline{underlined}.}
\label{tab:cross_kd_cifar100}
\resizebox{\textwidth}{!}{%
\begin{tabular}{
>{\raggedright\arraybackslash}m{2.2cm}
>{\raggedright\arraybackslash}m{1.8cm}
|cc|cccc|ccc|ccc>{\columncolor{gray!15}}c}
\toprule
\textbf{Teacher} & \textbf{Student} &
\multicolumn{2}{c|}{\textbf{From Scratch}} &
\multicolumn{4}{c|}{\textbf{Hint-based}} &
\multicolumn{3}{c|}{\textbf{Logits-based}} &
\multicolumn{4}{c}{\textbf{CAKD}} \\
\cmidrule(lr){3-4}\cmidrule(lr){5-8}\cmidrule(lr){9-11}\cmidrule(lr){12-15}
 & & T. & S. & FitNet & CC & RKD & CRD & KD & DKD & DIST & OFA & FBT & RSD & RPKD \\
\midrule
\multicolumn{15}{l}{\textit{CNN-based students}} \\
\midrule
Swin-T     & ResNet18    & 89.26 & 74.01 & 78.87 & 74.19 & 74.11 & 77.63 & 78.74 & 80.26 & 77.75 & 80.54 & 81.61 & \underline{83.92} & $\mathbf{85.15}_{\pm 0.18}$ \\
ViT-S      & ResNet18    & 92.04 & 74.01 & 77.71 & 74.26 & 73.72 & 76.60 & 77.26 & 78.10 & 76.49 & 80.15 & \underline{81.93} & 81.50 & $\mathbf{84.78}_{\pm 0.20}$ \\
Mixer-B/16 & ResNet18    & 87.29 & 74.01 & 77.15 & 74.26 & 73.75 & 76.42 & 77.79 & 78.67 & 76.36 & 79.39 & \textbf{81.90} & \underline{81.85} & $81.82_{\pm 0.21}$ \\
Swin-T     & MobileNetV2 & 89.26 & 73.68 & 74.28 & 71.19 & 69.00 & 79.80 & 74.68 & 71.07 & 72.89 & 80.98 & 81.28 & \underline{83.68} & $\mathbf{84.90}_{\pm 0.04}$ \\
ViT-S      & MobileNetV2 & 92.04 & 73.68 & 73.54 & 70.67 & 68.46 & 78.14 & 72.77 & 69.80 & 72.54 & 78.45 & \underline{82.10} & 81.68 & $\mathbf{84.17}_{\pm 0.11}$ \\
Mixer-B/16 & MobileNetV2 & 87.29 & 73.68 & 73.78 & 70.73 & 68.95 & 78.15 & 73.33 & 70.20 & 73.26 & 78.78 & 80.83 & \underline{81.74} & $\mathbf{82.43}_{\pm 0.11}$ \\
\midrule
\multicolumn{15}{l}{\textit{ViT-based students}} \\
\midrule
ConvNeXt-T & DeiT-T & 88.41 & 68.00 & 60.78 & 68.01 & 69.79 & 65.94 & 72.99 & 74.60 & 73.55 & 75.76 & 79.57 & \underline{82.46} & $\mathbf{83.74}_{\pm 0.25}$ \\
Mixer-B/16 & DeiT-T & 87.29 & 68.00 & 71.05 & 68.13 & 69.89 & 65.35 & 71.36 & 73.44 & 71.67 & 73.90 & 74.40 & \textbf{78.50} & $\underline{76.66}_{\pm 0.05}$ \\
ConvNeXt-T & Swin-P & 88.41 & 72.63 & 24.06 & 72.63 & 71.73 & 67.09 & 76.44 & 76.80 & 76.41 & 78.32 & \underline{80.73} & \textbf{82.21} & $79.98_{\pm 0.26}$ \\
Mixer-B/16 & Swin-P & 87.29 & 72.63 & 75.20 & 73.32 & 70.82 & 67.03 & 75.93 & 76.39 & 75.85 & \underline{78.93} & 78.44 & \textbf{81.28} & $75.50_{\pm 0.07}$ \\
\bottomrule
\end{tabular}%
}
\end{table}

\paragraph{CIFAR-100 cross-architecture results.}
Table~\ref{tab:cross_kd_cifar100} reports top-1 accuracy across all 10 pairs.
RPKD achieves a maximum gain of $\mathbf{+7.98\%}$ over OFA.

Among CNN-based students, RPKD achieves $85.15\%$ (Swin-T $\to$ ResNet18) and $84.90\%$ (Swin-T $\to$ MobileNetV2), corresponding to gains of $+4.61\%$ and $+3.92\%$ over OFA.
The fragility of direct feature matching under heterogeneous architectures is illustrated by FitNet, which collapses to $24.06\%$ on ConvNeXt-T $\to$ Swin-P when feature spaces are incompatible.

Among ViT-based students, RPKD leads on ConvNeXt-T $\to$ DeiT-T ($83.74\%$, $+7.98$ pp over OFA and $+1.28$ pp over RSD~\cite{zhang2025cross}), where the CNN-to-ViT structural gap is greatest.
On the remaining three ViT-student pairs, RSD leads; all three involve an MLP-Mixer teacher or the Swin-P student, where token-mixing weakens the spatial structure that the RPKD feature branch relies on.

\begin{table}[t]
\centering
\setlength{\tabcolsep}{1pt}
\small
\caption{Cross-architecture KD results on ImageNet-100. OFA, RSD and RPKD are our own runs, each averaged over three seeds; standard deviations are shown for RPKD. The remaining baselines are single-run. Best results in \textbf{bold}, second best \underline{underlined}.}
\label{tab:cross_kd_imagenet100}
\resizebox{\textwidth}{!}{%
\begin{tabular}{
>{\raggedright\arraybackslash}m{2.2cm}
>{\raggedright\arraybackslash}m{1.8cm}
|cc|cccc|ccc|cc>{\columncolor{gray!15}}c}
\toprule
\textbf{Teacher} & \textbf{Student} &
\multicolumn{2}{c|}{\textbf{From Scratch}} &
\multicolumn{4}{c|}{\textbf{Hint-based}} &
\multicolumn{3}{c|}{\textbf{Logits-based}} &
\multicolumn{3}{c}{\textbf{CAKD}} \\
\cmidrule(lr){3-4}\cmidrule(lr){5-8}\cmidrule(lr){9-11}\cmidrule(lr){12-14}
 & & T. & S. & FitNet & CC & RKD & CRD & KD & DKD & DIST & OFA & RSD & RPKD \\
\midrule
\multicolumn{14}{l}{\textit{CNN-based students}} \\
\midrule
DeiT-T     & ResNet18    & 89.72 & 81.58 & 85.84 & 81.70 & 83.14 & 84.54 & 84.32 & 85.50 & 84.52 & 85.46 & \underline{87.14} & $\mathbf{87.87}_{\pm 0.24}$ \\
Swin-T     & ResNet18    & 94.48 & 81.58 & 81.94 & 81.68 & 82.78 & 83.44 & 82.76 & 84.70 & 83.42 & 84.67 & \underline{87.25} & $\mathbf{87.86}_{\pm 0.33}$ \\
Mixer-B/16 & ResNet18    & 90.42 & 81.58 & 82.58 & 81.98 & 81.90 & 83.88 & 82.84 & 84.88 & 83.38 & 84.19 & \textbf{85.85} & $\underline{85.37}_{\pm 0.20}$ \\
DeiT-T     & MobileNetV2 & 89.72 & 83.51 & 86.94 & 84.24 & 83.94 & 85.06 & 86.14 & 86.44 & 85.38 & 85.99 & \underline{87.01} & $\mathbf{87.35}_{\pm 0.08}$ \\
Swin-T     & MobileNetV2 & 94.48 & 83.51 & 85.20 & 83.84 & 82.94 & 84.92 & 84.96 & 86.18 & 84.80 & 85.24 & \textbf{87.17} & $\underline{86.95}_{\pm 0.18}$ \\
Mixer-B/16 & MobileNetV2 & 90.42 & 83.51 & 84.78 & 84.06 & 83.28 & 84.12 & 83.90 & \underline{85.62} & 84.62 & 85.06 & \textbf{86.06} & $85.07_{\pm 0.39}$ \\
\midrule
\multicolumn{14}{l}{\textit{ViT-based students}} \\
\midrule
ResNet50   & DeiT-T & 93.52 & 63.88 & 64.62 & 63.68 & 64.62 & 56.94 & 65.76 & 66.24 & 65.34 & 68.56 & \underline{72.18} & $\mathbf{81.00}_{\pm 0.12}$ \\
ConvNeXt-T & DeiT-T & 94.76 & 63.88 & 63.38 & 64.24 & 63.64 & 58.78 & 66.08 & 63.42 & 64.82 & 67.38 & \underline{67.93} & $\mathbf{79.14}_{\pm 0.26}$ \\
Mixer-B/16 & DeiT-T & 90.42 & 63.88 & 65.96 & 64.66 & 65.22 & 55.30 & 68.12 & 66.88 & 67.52 & \underline{69.75} & 69.21 & $\mathbf{70.20}_{\pm 0.04}$ \\
ResNet50   & Swin-N & 93.52 & 78.78 & 79.16 & 78.90 & 75.30 & 75.24 & 80.94 & 80.62 & 80.50 & 80.38 & \underline{84.31} & $\mathbf{84.57}_{\pm 0.16}$ \\
ConvNeXt-T & Swin-N & 94.76 & 78.78 & 78.90 & 78.82 & 73.66 & 75.44 & 79.90 & 79.86 & 80.60 & 79.93 & \underline{83.09} & $\mathbf{83.15}_{\pm 0.21}$ \\
Mixer-B/16 & Swin-N & 90.42 & 78.78 & 79.52 & 78.76 & 74.02 & 74.62 & 81.28 & \underline{81.40} & 80.90 & 81.19 & \textbf{82.57} & $79.92_{\pm 0.13}$ \\
\bottomrule
\end{tabular}%
}
\end{table}

\paragraph{ImageNet-100 cross-architecture results.}
Table~\ref{tab:cross_kd_imagenet100} extends the evaluation to 12 teacher--student pairs on ImageNet-100.
RPKD ranks first on 8 of 12 pairs and exceeds OFA on 11 of 12, improving over the student baseline on every pair; the four pairs it does not lead are all won by RSD~\cite{zhang2025cross}.
The gains are most pronounced for ViT-based students paired with CNN teachers: RPKD reaches $81.00\%$ on ResNet50 $\to$ DeiT-T ($+12.44\%$ over OFA) and $79.14\%$ on ConvNeXt-T $\to$ DeiT-T ($+11.76\%$ over OFA), suggesting that prototype-based transfer succeeds precisely by avoiding model-specific structural assumptions, relying instead on class-similarity geometry that is approximately preserved under random projection regardless of architectural boundaries.
CNN-based students show consistent but more moderate improvements, with RPKD outperforming OFA on all six CNN-student pairs.
The single exception is Mixer-B/16 $\to$ Swin-N, where RSD ($82.57\%$) outperforms RPKD ($79.92\%$), consistent with the analogous Mixer-B/16 $\to$ Swin-P result on CIFAR-100.

\paragraph{MLP-Mixer teacher behavior.}
When MLP-Mixer acts as teacher, RPKD underperforms relative to CNN and ViT teacher settings and in several pairs fails to outperform OFA --- a trend shared by all feature-based methods across both benchmarks.
We attribute this to the token-mixing nature of MLP-Mixers: repeated spatial mixing weakens localized positional correspondence in the representations, undermining the spatial alignment that the feature branch relies on.

\paragraph{ImageNet-1K cross-architecture results.}
Table~\ref{tab:cross_kd_imagenet} reports full-scale ImageNet-1K results for four cross-architecture pairs.
RPKD exceeds OFA on all four pairs and leads overall on DeiT-T $\to$ MobileNetV2 ($72.67\%$, $+1.28\%$ over OFA and $+0.49\%$ over RSD~\cite{zhang2025cross}).
On the remaining pairs RPKD stays within $0.32$ pp of the better of FBT~\cite{li2025fuse} and RSD, indicating that at full ImageNet-1K scale the three CAKD methods are broadly comparable.
The larger gain on MobileNetV2 is consistent with the CIFAR-100 trend: the narrow channel bottlenecks of inverted-residual architectures~\cite{sandler2018mobilenetv2} produce low-rank features that prototype-space supervision handles better than direct learned alignment.

\begin{table}[t]
\centering
\setlength{\tabcolsep}{2pt}
\small
\caption{Cross-architecture KD results on ImageNet-1K. Best results in \textbf{bold}, second best \underline{underlined}.}
\label{tab:cross_kd_imagenet}
\resizebox{\textwidth}{!}{%
\begin{tabular}{ll|cc|cccc|ccc|ccc>{\columncolor{gray!15}}c}
\toprule
\textbf{Teacher} & \textbf{Student} &
\multicolumn{2}{c|}{\textbf{From Scratch}} &
\multicolumn{4}{c|}{\textbf{Hint-based}} &
\multicolumn{3}{c|}{\textbf{Logits-based}} &
\multicolumn{4}{c}{\textbf{CAKD}} \\
\cmidrule(lr){3-4}\cmidrule(lr){5-8}\cmidrule(lr){9-11}\cmidrule(lr){12-15}
 & & T. & S. & FitNet & CC & RKD & CRD & KD & DKD & DIST & OFA & FBT & RSD & RPKD \\
\midrule
DeiT-T & ResNet18    & 72.17 & 69.75 & 70.44 & 69.77 & 69.47 & 69.25 & 70.22 & 69.39 & 70.64 & 71.34 & 71.22 & \textbf{71.70} & $\underline{71.38}_{\pm 0.14}$ \\
Swin-T & ResNet18    & 81.38 & 69.75 & 71.18 & 70.07 & 68.89 & 69.09 & 71.14 & 71.10 & 70.91 & 71.85 & \textbf{72.21} & \underline{72.13} & $72.01_{\pm 0.12}$ \\
DeiT-T & MobileNetV2 & 72.17 & 68.87 & 70.95 & 70.69 & 69.72 & 69.60 & 70.87 & 70.14 & 71.08 & 71.39 & 71.78 & \underline{72.18} & $\mathbf{72.67}_{\pm 0.11}$ \\
Swin-T & MobileNetV2 & 81.38 & 68.87 & 71.75 & 70.69 & 67.52 & 69.58 & 72.05 & 71.71 & 71.76 & 72.32 & \textbf{72.54} & 72.36 & $\underline{72.40}_{\pm 0.21}$ \\
\bottomrule
\end{tabular}%
}
\end{table}

\paragraph{Homogeneous architecture results.}
Table~\ref{tab:kd_homo} evaluates RPKD in the standard same-family setting on ImageNet-1K, where learned alignment methods operate at their best.
On ResNet34 $\to$ ResNet18, RPKD reaches $72.23\%$, ahead of OFA ($72.10\%$) and RSD~\cite{zhang2025cross} ($72.18\%$) and behind FBT~\cite{li2025fuse} ($72.29\%$); the three leading methods span only $0.11$ pp, less than RPKD's own run-to-run variation ($\pm 0.13$), so this pair is better read as a tie among recent methods than as a ranking.
On ResNet50 $\to$ MobileNetV1 the margin is unambiguous: RPKD achieves $74.20\%$, surpassing every prior method including FBT ($73.45\%$) by $+0.75$ pp.
Prototype-space distillation therefore generalizes to standard CNN-to-CNN settings, matching learned-alignment methods where they are strongest and exceeding them when the student is narrow.

\begin{table}[t]
\centering
\setlength{\tabcolsep}{4pt}
\small
\caption{Standard CNN-to-CNN KD on ImageNet-1K. Best in \textbf{bold}, second best \underline{underlined}.}
\label{tab:kd_homo}
\resizebox{\textwidth}{!}{%
\begin{tabular}{ll|cc|cccccc|ccc>{\columncolor{gray!15}}c}
\toprule
\textbf{Teacher} & \textbf{Student} & T. & S. & KD & OFD & Review & CRD & DKD & DIST & OFA & FBT & RSD & RPKD \\
\midrule
ResNet34 & ResNet18    & 73.31 & 69.75 & 70.66 & 70.81 & 71.61 & 71.17 & 71.70 & 72.07 & 72.10 & \textbf{72.29} & 72.18 & $\underline{72.23}_{\pm 0.13}$ \\
ResNet50 & MobileNetV1 & 76.16 & 68.58 & 68.58 & 71.25 & 72.56 & 71.37 & 72.05 & 73.24 & 73.24 & \underline{73.45} & 73.08 & $\mathbf{74.20}_{\pm 0.09}$ \\
\bottomrule
\end{tabular}}
\end{table}

\paragraph{Low-category advantage.}
A distinctive property of RPKD is that the prototype vocabulary is decoupled from the number of task classes: the student always receives $M\!=\!500$-dimensional supervisory signal regardless of dataset size, whereas logit-based soft targets collapse toward near-uniform distributions as the class count decreases.
Table~\ref{tab:in_subset} tests this on ImageNet subsets of 10, 20, 50 and 100 classes (DeiT-T $\to$ ResNet18), formed by taking the first 10/20/50/100 classes of ImageNet-100~\cite{tian2020contrastive} ordered alphabetically by synset name.
At IN-10, both hint-based CRD ($88.60$) and logit-based DIST ($89.80$) fall \emph{below} the student trained from scratch ($90.80$), while RPKD reaches $95.40$, a $+4.60\%$ gain over the student and $+2.80\%$ over OFA; the advantage holds at every subset size, with $+3.00\%$ over OFA on IN-20 and $+2.41\%$ on IN-100.
The scaling is what the design predicts: as the class count drops from 100 to 10, KD's gain over the student shrinks from $+2.74$ to $+0.60$ pp and OFA's from $+3.88$ to $+1.80$ pp, while RPKD's stays between $+6.29$ and $+4.60$ pp, so its advantage over OFA widens from roughly $1.5$--$1.6\times$ at 50--100 classes to $2.3$--$2.6\times$ at 10--20.
Each branch is individually competitive --- $\text{RPKD}_{\text{feat}}$ ranks second at IN-10 and IN-20, $\text{RPKD}_{\text{logit}}$ at IN-50 and IN-100 --- yet the full method outperforms both at every size, confirming that the two objectives are complementary.

\begin{table}[!ht]
\centering
\setlength{\tabcolsep}{3pt}
\small
\caption{Top-1 accuracy on ImageNet subsets of varying class count (T: DeiT-T, S: ResNet18). IN-10/20/50 use the first 10/20/50 ImageNet-100~\cite{tian2020contrastive} classes and are single run; in the IN-100 row the OFA and RPKD entries repeat the three-seed means of Table~\ref{tab:cross_kd_imagenet100}. Best in \textbf{bold}, second best \underline{underlined}.}
\label{tab:in_subset}
\resizebox{\textwidth}{!}{%
\begin{tabular}{l|cc|cccc|cccc|cc>{\columncolor{gray!15}}c}
\toprule
\textbf{Subset} &
\multicolumn{2}{c|}{\textbf{From Scratch}} &
\multicolumn{4}{c|}{\textbf{Hint-based}} &
\multicolumn{4}{c|}{\textbf{Logits-based}} &
\multicolumn{3}{c}{\textbf{Ours}} \\
\cmidrule(lr){2-3}\cmidrule(lr){4-7}\cmidrule(lr){8-11}\cmidrule(lr){12-14}
 & T. & S. & FitNet & CC & RKD & CRD & KD & DKD & DIST & OFA &
$\text{RPKD}_{\text{logit}}$ & $\text{RPKD}_{\text{feat}}$ & RPKD \\
\midrule
IN-10  & 97.80 & 90.80 & 93.20 & 91.80 & 92.80 & 88.60 & 91.40 & 91.80 & 89.80 & 92.60 & 93.80 & \underline{95.20} & \textbf{95.40} \\
IN-20  & 97.40 & 89.80 & 92.80 & 90.00 & 90.20 & 88.10 & 90.80 & 92.20 & 90.60 & 92.20 & 92.30 & \underline{94.80} & \textbf{95.20} \\
IN-50  & 93.28 & 84.76 & 88.80 & 83.60 & 85.84 & 86.12 & 87.20 & 88.32 & 86.68 & 88.48 & \underline{88.88} & 88.80 & \textbf{90.44} \\
IN-100 & 89.72 & 81.58 & 85.84 & 81.70 & 83.14 & 84.54 & 84.32 & 85.50 & 84.52 & 85.46 & \underline{86.22} & 86.16 & \textbf{87.87} \\
\bottomrule
\end{tabular}%
}
\end{table}

\paragraph{Ablation: feature extraction stage.}
\begin{table}[t]
\centering
\small
\caption{Ablation of feature extraction stage on IN-100 (DeiT-T $\to$ ResNet18). Stages 1--4 denote depth from shallow to deep; \{4\} is the default. Single run, except \{4\} (three-seed mean, cf.\ Table~\ref{tab:cross_kd_imagenet100}).}
\label{tab:ablation_stage}
\setlength{\tabcolsep}{14pt}
\begin{tabular}{lcccc}
\toprule
\textbf{Stages used} & \{1, 2, 3, 4\} & \{2, 3, 4\} & \{3, 4\} & \{4\} (default) \\
\midrule
\textbf{Top-1 Acc.} & 86.30 & 86.80 & 87.65 & \textbf{87.87} \\
\bottomrule
\end{tabular}
\end{table}
Table~\ref{tab:ablation_stage} ablates the network stages used for feature extraction, evaluated on IN-100 (DeiT-T $\to$ ResNet18).
Each network is divided into four stages of increasing depth, and distillation is progressively restricted to later stages only.
Performance rises monotonically as earlier stages are removed: using all four stages yields $86.30\%$, while the final stage alone achieves $87.87\%$.
Early-stage features encode low-level spatial statistics that are hard to reconcile across heterogeneous architectures through prototype matching; final-stage features (as in all other experiments) carry high-level semantics that align more readily in the shared prototype space.

\paragraph{Ablation: prototype count and dimension.}
Table~\ref{tab:ablation_mr} ablates the prototype count $M$ and embedding dimension $R$ on IN-100 over three independent seeds, for a CNN student (DeiT-T $\to$ ResNet18) and a ViT student (ResNet50 $\to$ DeiT-T).
Across a $16\times$ sweep of either hyperparameter the total range is $0.22$ and $0.27$ pp for the CNN student---within its training noise ($\pm 0.24$ pp)---and $0.84$ and $1.27$ pp for the ViT student, whose only notable preference is $R\!=\!512$.
RPKD retains its margin over OFA at every setting ($\geq\!+2.14$ pp and $\geq\!+11.80$ pp respectively), so the defaults were fixed once and reused across all 28 pairs without per-pair tuning.
The plateau around $M\!=\!500$ is consistent with the Johnson--Lindenstrauss argument: for batch size $n\!=\!128$ and distortion $\varepsilon\!=\!0.1$, a sufficient threshold is $M \gtrsim \ln(128)/0.1^2 \approx 485$, placing our default near this bound; beyond it, additional prototypes yield diminishing returns.
Appendix~\ref{sec:appendix_proto_seed} further shows that varying only the random draw of $\mathbf{P}$ moves accuracy no more than varying the training seed.

\begin{table}[t]
\centering
\small
\caption{Ablation of prototype count $M$ (left, $R\!=\!256$) and embedding dimension $R$ (right, $M\!=\!500$) on IN-100. Mean $\pm$ std over 3 seeds, using an independent seed set from Table~\ref{tab:cross_kd_imagenet100}. Defaults in \textbf{bold}.}
\label{tab:ablation_mr}
\setlength{\tabcolsep}{6pt}
\begin{tabular}{ccc@{\hskip 24pt}ccc}
\toprule
\multicolumn{3}{c}{\textbf{Prototype count} $M$ ($R\!=\!256$)} & \multicolumn{3}{c}{\textbf{Embedding dimension} $R$ ($M\!=\!500$)} \\
\cmidrule(lr){1-3}\cmidrule(lr){4-6}
$M$ & DeiT-T $\to$ RN18 & RN50 $\to$ DeiT-T & $R$ & DeiT-T $\to$ RN18 & RN50 $\to$ DeiT-T \\
\midrule
125  & $87.65_{\pm 0.13}$ & $80.36_{\pm 0.10}$ & 64   & $87.60_{\pm 0.09}$ & $80.63_{\pm 0.45}$ \\
250  & $87.74_{\pm 0.16}$ & $81.07_{\pm 0.15}$ & 128  & $87.72_{\pm 0.14}$ & $80.93_{\pm 0.17}$ \\
\textbf{500} & $\mathbf{87.87_{\pm 0.24}}$ & $\mathbf{81.20_{\pm 0.29}}$ & \textbf{256} & $\mathbf{87.87_{\pm 0.24}}$ & $81.20_{\pm 0.29}$ \\
1000 & $87.67_{\pm 0.32}$ & $81.01_{\pm 0.46}$ & 512  & $87.71_{\pm 0.19}$ & $\mathbf{81.80_{\pm 0.27}}$ \\
2000 & $87.79_{\pm 0.28}$ & $81.13_{\pm 0.06}$ & 1024 & $87.63_{\pm 0.22}$ & $80.53_{\pm 0.19}$ \\
\midrule
\textit{Range} & $0.22$ & $0.84$ & \textit{Range} & $0.27$ & $1.27$ \\
\bottomrule
\end{tabular}
\end{table}

\paragraph{Ablation: frozen vs.\ learned prototypes.}
Table~\ref{tab:frozen_learned} tests the design choice at the core of RPKD: whether the prototype bank $\mathbf{P}$ should be frozen at initialization or learned jointly with the student projection heads.
We run four architecturally diverse IN-100 pairs at $M\!=\!R\!=\!500$ (so that exact orthogonality is available for the Frozen Orthogonal variant) over three independent seeds, holding all other components and hyperparameters fixed.
Every frozen variant outperforms the learned prototype on every pair, by $+0.57$ pp (DeiT-T $\to$ RN18), $+2.86$ pp (ConvNeXt-T $\to$ Swin-N), $+3.84$ pp (RN50 $\to$ DeiT-T), and $+0.89$ pp (Swin-T $\to$ MBv2): a fixed random geometry consistently beats an optimized, data-adapted one, and the margin grows for heterogeneous pairs.
Among frozen variants no single structure dominates (maximum pairwise difference $0.54$ pp), which indicates that what matters is a \emph{fixed} reference geometry rather than the specific basis.
A learned $\mathbf{P}$ co-evolves with the student projection heads and therefore presents a moving target; because both $\mathbf{P}$ and the teacher projections are frozen, the student instead aligns against a completely static basis throughout training.

\begin{table}[t]
\centering
\small
\caption{Frozen vs.\ learned prototype ablation on ImageNet-100 (mean $\pm$ std, 3 seeds; $M\!=\!R\!=\!500$). Frozen Gaussian uses $\ell_2$-normalized random rows, Frozen Orthogonal a QR-decomposed square matrix, and Frozen DCT the cosine frequency basis of $\mathbb{R}^{500}$. Bold = best per column.}
\label{tab:frozen_learned}
\resizebox{\textwidth}{!}{%
\setlength{\tabcolsep}{6pt}
\begin{tabular}{lcccc}
\toprule
\textbf{Prototype} & \textbf{DeiT-T $\to$ RN18} & \textbf{ConvNeXt-T $\to$ Swin-N} & \textbf{RN50 $\to$ DeiT-T} & \textbf{Swin-T $\to$ MBv2} \\
\midrule
Frozen Gaussian    & $87.78_{\pm 0.29}$ & $\mathbf{83.90_{\pm 0.24}}$ & $80.93_{\pm 0.38}$ & $\mathbf{87.10_{\pm 0.16}}$ \\
Frozen Orthogonal  & $\mathbf{87.86_{\pm 0.27}}$ & $83.47_{\pm 0.12}$ & $80.67_{\pm 0.36}$ & $87.09_{\pm 0.01}$ \\
Frozen DCT         & $87.71_{\pm 0.06}$ & $83.73_{\pm 0.18}$ & $\mathbf{81.21_{\pm 0.29}}$ & $87.01_{\pm 0.30}$ \\
\midrule
Learned (Gaussian-init) & $87.29_{\pm 0.10}$ & $81.04_{\pm 0.03}$ & $77.37_{\pm 0.38}$ & $86.21_{\pm 0.15}$ \\
\bottomrule
\end{tabular}%
}
\end{table}

\paragraph{Ablation: branch contributions.}
Appendix~\ref{sec:appendix_component} isolates the two branches on six ImageNet-100 and five ImageNet-1K pairs.
Each branch independently yields substantial gains over the CE-only baseline, and their relative importance is architecture-dependent: for CNN $\to$ ViT distillation the spatial branch $\mathcal{L}_{\mathrm{feat}}$ does most of the work in bridging the representation gap, whereas for MLP-Mixer teachers, where token-mixing weakens spatial structure, the global branch drives performance.
Combined, the two are complementary: on ImageNet-1K full RPKD strictly outperforms both individual branches on all five pairs.

\paragraph{Limitations.}
RPKD's performance is robust to the prototype count $M$ and embedding dimension $R$ (total range $\leq 0.27$ pp across $16\times$ sweeps), and the distillation temperature is fixed at $T\!=\!4.0$ throughout, but the loss coefficients require a small grid search; adaptive weighting schemes could reduce this tuning burden.
Its effectiveness is also bounded by teacher quality: a poorly calibrated or biased teacher transfers suboptimal representations.
While ImageNet-100 provides 12-pair coverage across CNN and ViT families, full-scale ImageNet-1K evaluation is restricted to four CNN-based student pairs owing to computational constraints.
RPKD has been evaluated exclusively on image classification; its applicability to other recognition problems such as object detection and semantic segmentation requires further study.
Finally, RPKD's effectiveness degrades when the teacher is an MLP-Mixer, as the token-mixing nature of such architectures weakens the spatial correspondence that the feature branch relies on.

\section{Conclusion}
\label{sec:conclusion}

We presented \textbf{RPKD}, a knowledge distillation framework built on the observation that effective cross-architecture knowledge transfer does not require learned feature alignment.
By projecting both teacher and student into a shared random prototype space that is fixed at initialization and reused without modification across all teacher--student pairs, RPKD removes the need for a learned alignment space, for joint teacher--student optimization, and for any pair-specific change to the objective.
Knowledge is transferred through two complementary objectives: a logit branch that aligns global prototype similarity distributions and a feature branch that matches spatially-resolved activation maps, jointly encoding semantic and structural information within the same frozen space.

\textbf{Empirical results.}
On CIFAR-100, across 10 cross-architecture pairs spanning convolutional, transformer, and MLP families, RPKD improves over training from scratch on every pair, exceeding OFA by up to $\mathbf{+7.98\%}$ and reaching $85.15\%$ on Swin-T $\to$ ResNet18.
On ImageNet-100 it ranks first on 8 of 12 pairs and exceeds OFA on 11 of 12, with a maximum gain of $\mathbf{+12.44\%}$ over OFA.
On ImageNet-1K it exceeds OFA by up to $+1.28\%$ on cross-architecture pairs, ties the leading recent methods on ResNet34 $\to$ ResNet18 ($72.23\%$, within run-to-run variation), and reaches $74.20\%$ on ResNet50 $\to$ MobileNetV1, ahead of all prior methods by at least $+0.75$ pp.
The low-category advantage is pronounced: on 10- and 20-class subsets RPKD surpasses OFA by $+2.80\%$ and $+3.00\%$, while hint- and logit-based methods degrade as the label space shrinks.

\textbf{Implications.}
That a random, frozen prototype space suffices as a distillation medium challenges the prevailing assumption that cross-architecture alignment must be learned: the information bottleneck of a random projection is enough to align teacher and student similarity structures across large architectural gaps.
The entire teacher side reduces to two frozen random matrices with no trainable parameters, yet matches or outperforms learned alignment while training faster and with no more peak memory (Appendix~\ref{sec:appendix_efficiency}).
We hope RPKD encourages further study of structure-preserving random projections as knowledge transfer media.

\newpage
\label{sec:ackstart}
\begin{ack}
We are grateful to Dr. Mert Bülent Sarıyıldız for his valuable guidance on the knowledge distillation literature during the early stages of this work. The numerical calculations reported in this paper were performed on MareNostrum~5, operated by the Barcelona Supercomputing Center (BSC-CNS), through the TURMETUGC project supported by T\"UB\.ITAK\@.
This study is supported in part by the ADEP project with grant no. ADEP-312-2024-11525. G. Cinbis is supported by the “Young Scientist Awards Program (BAGEP)” of Science Academy, Türkiye. The authors declare no competing interests. 
\end{ack}

\bibliographystyle{unsrtnat}
\bibliography{references}

\begin{thebibliography}{50}
\providecommand{\natexlab}[1]{#1}
\providecommand{\url}[1]{\texttt{#1}}
\expandafter\ifx\csname urlstyle\endcsname\relax
  \providecommand{\doi}[1]{doi: #1}\else
  \providecommand{\doi}{doi: \begingroup \urlstyle{rm}\Url}\fi

\bibitem[Hinton et~al.(2015)Hinton, Vinyals, and Dean]{hinton2015distilling}
Geoffrey Hinton, Oriol Vinyals, and Jeff Dean.
\newblock Distilling the knowledge in a neural network.
\newblock \emph{arXiv preprint arXiv:1503.02531}, 2015.

\bibitem[Gou et~al.(2021)Gou, Yu, Maybank, and Tao]{gou2021knowledge}
Jianping Gou, Baosheng Yu, Stephen~J Maybank, and Dacheng Tao.
\newblock Knowledge distillation: A survey.
\newblock \emph{International journal of computer vision}, 129\penalty0
  (6):\penalty0 1789--1819, 2021.

\bibitem[Furlanello et~al.(2018)Furlanello, Lipton, Tschannen, Itti, and
  Anandkumar]{furlanello2018born}
Tommaso Furlanello, Zachary Lipton, Michael Tschannen, Laurent Itti, and Anima
  Anandkumar.
\newblock Born again neural networks.
\newblock In \emph{International conference on machine learning}, pages
  1607--1616. PMLR, 2018.

\bibitem[Zhang et~al.(2018)Zhang, Xiang, Hospedales, and Lu]{zhang2018deep}
Ying Zhang, Tao Xiang, Timothy~M Hospedales, and Huchuan Lu.
\newblock Deep mutual learning.
\newblock In \emph{Proceedings of the IEEE conference on computer vision and
  pattern recognition}, pages 4320--4328, 2018.

\bibitem[Jin et~al.(2019)Jin, Peng, Wu, Liu, Liu, Liang, Yan, and
  Hu]{jin2019knowledge}
Xiao Jin, Baoyun Peng, Yichao Wu, Yu~Liu, Jiaheng Liu, Ding Liang, Junjie Yan,
  and Xiaolin Hu.
\newblock Knowledge distillation via route constrained optimization.
\newblock In \emph{Proceedings of the IEEE/CVF international conference on
  computer vision}, pages 1345--1354, 2019.

\bibitem[Son et~al.(2021)Son, Na, Choi, and Hwang]{son2021densely}
Wonchul Son, Jaemin Na, Junyong Choi, and Wonjun Hwang.
\newblock Densely guided knowledge distillation using multiple teacher
  assistants.
\newblock In \emph{Proceedings of the IEEE/CVF international conference on
  computer vision}, pages 9395--9404, 2021.

\bibitem[Huang et~al.(2022)Huang, You, Wang, Qian, and Xu]{huang2022knowledge}
Tao Huang, Shan You, Fei Wang, Chen Qian, and Chang Xu.
\newblock Knowledge distillation from a stronger teacher.
\newblock \emph{Advances in neural information processing systems},
  35:\penalty0 33716--33727, 2022.

\bibitem[Zhao et~al.(2022)Zhao, Cui, Song, Qiu, and Liang]{zhao2022decoupled}
Borui Zhao, Quan Cui, Renjie Song, Yiyu Qiu, and Jiajun Liang.
\newblock Decoupled knowledge distillation.
\newblock In \emph{Proceedings of the IEEE/CVF Conference on computer vision
  and pattern recognition}, pages 11953--11962, 2022.

\bibitem[Sun et~al.(2024)Sun, Ren, Li, Wang, and Cao]{sun2024logit}
Shangquan Sun, Wenqi Ren, Jingzhi Li, Rui Wang, and Xiaochun Cao.
\newblock Logit standardization in knowledge distillation.
\newblock In \emph{Proceedings of the IEEE/CVF conference on computer vision
  and pattern recognition}, pages 15731--15740, 2024.

\bibitem[Romero et~al.(2015)Romero, Ballas, Kahou, Chassang, Gatta, and
  Bengio]{romero2015fitnets}
Adriana Romero, Nicolas Ballas, Samira~Ebrahimi Kahou, Antoine Chassang, Carlo
  Gatta, and Yoshua Bengio.
\newblock Fitnets: Hints for thin deep nets.
\newblock In \emph{International Conference on Learning Representations
  (ICLR)}, 2015.

\bibitem[Zagoruyko and Komodakis(2016)]{zagoruyko2016paying}
Sergey Zagoruyko and Nikos Komodakis.
\newblock Paying more attention to attention: Improving the performance of
  convolutional neural networks via attention transfer.
\newblock \emph{arXiv preprint arXiv:1612.03928}, 2016.

\bibitem[Yim et~al.(2017)Yim, Joo, Bae, and Kim]{yim2017gift}
Junho Yim, Donggyu Joo, Jihoon Bae, and Junmo Kim.
\newblock A gift from knowledge distillation: Fast optimization, network
  minimization and transfer learning.
\newblock In \emph{Proceedings of the IEEE conference on computer vision and
  pattern recognition}, pages 4133--4141, 2017.

\bibitem[Heo et~al.(2019{\natexlab{a}})Heo, Lee, Yun, and
  Choi]{heo2019knowledge}
Byeongho Heo, Minsik Lee, Sangdoo Yun, and Jin~Young Choi.
\newblock Knowledge transfer via distillation of activation boundaries formed
  by hidden neurons.
\newblock In \emph{Proceedings of the AAAI conference on artificial
  intelligence}, volume~33, pages 3779--3787, 2019{\natexlab{a}}.

\bibitem[Heo et~al.(2019{\natexlab{b}})Heo, Kim, Yun, Park, Kwak, and
  Choi]{heo2019comprehensive}
Byeongho Heo, Jeesoo Kim, Sangdoo Yun, Hyojin Park, Nojun Kwak, and Jin~Young
  Choi.
\newblock A comprehensive overhaul of feature distillation.
\newblock In \emph{Proceedings of the IEEE/CVF international conference on
  computer vision}, pages 1921--1930, 2019{\natexlab{b}}.

\bibitem[Huang and Wang(2017)]{huang2017like}
Zehao Huang and Naiyan Wang.
\newblock Like what you like: Knowledge distill via neuron selectivity
  transfer.
\newblock \emph{arXiv preprint arXiv:1707.01219}, 2017.

\bibitem[Peng et~al.(2019)Peng, Jin, Liu, Li, Wu, Liu, Zhou, and
  Zhang]{peng2019correlation}
Baoyun Peng, Xiao Jin, Jiaheng Liu, Dongsheng Li, Yichao Wu, Yu~Liu, Shunfeng
  Zhou, and Zhaoning Zhang.
\newblock Correlation congruence for knowledge distillation.
\newblock In \emph{Proceedings of the IEEE/CVF international conference on
  computer vision}, pages 5007--5016, 2019.

\bibitem[Park et~al.(2019)Park, Kim, Lu, and Cho]{park2019relational}
Wonpyo Park, Dongju Kim, Yan Lu, and Minsu Cho.
\newblock Relational knowledge distillation.
\newblock In \emph{Proceedings of the IEEE/CVF conference on computer vision
  and pattern recognition}, pages 3967--3976, 2019.

\bibitem[Liu et~al.(2019)Liu, Cao, Li, Yuan, Hu, Li, and
  Duan]{liu2019knowledge}
Yufan Liu, Jiajiong Cao, Bing Li, Chunfeng Yuan, Weiming Hu, Yangxi Li, and
  Yunqiang Duan.
\newblock Knowledge distillation via instance relationship graph.
\newblock In \emph{Proceedings of the IEEE/CVF conference on computer vision
  and pattern recognition}, pages 7096--7104, 2019.

\bibitem[Tian et~al.(2019)Tian, Krishnan, and Isola]{tian2019contrastive}
Yonglong Tian, Dilip Krishnan, and Phillip Isola.
\newblock Contrastive representation distillation.
\newblock \emph{arXiv preprint arXiv:1910.10699}, 2019.

\bibitem[Ahn et~al.(2019)Ahn, Hu, Damianou, Lawrence, and
  Dai]{ahn2019variational}
Sungsoo Ahn, Shell~Xu Hu, Andreas Damianou, Neil~D Lawrence, and Zhenwen Dai.
\newblock Variational information distillation for knowledge transfer.
\newblock In \emph{Proceedings of the IEEE/CVF conference on computer vision
  and pattern recognition}, pages 9163--9171, 2019.

\bibitem[Lin et~al.(2022)Lin, Xie, Wang, Yu, Chang, Liang, and
  Wang]{lin2022knowledge}
Sihao Lin, Hongwei Xie, Bing Wang, Kaicheng Yu, Xiaojun Chang, Xiaodan Liang,
  and Gang Wang.
\newblock Knowledge distillation via the target-aware transformer.
\newblock In \emph{Proceedings of the IEEE/CVF conference on computer vision
  and pattern recognition}, pages 10915--10924, 2022.

\bibitem[Chen et~al.(2022{\natexlab{a}})Chen, Mei, Zhang, Wang, Feng, and
  Chen]{chen2022knowledge}
Defang Chen, Jian-Ping Mei, Hailin Zhang, Can Wang, Yan Feng, and Chun Chen.
\newblock Knowledge distillation with the reused teacher classifier.
\newblock In \emph{Proceedings of the IEEE/CVF conference on computer vision
  and pattern recognition}, pages 11933--11942, 2022{\natexlab{a}}.

\bibitem[Zhang et~al.(2020)Zhang, Shi, Shi, Ma, and Bao]{zhang2020task}
Linfeng Zhang, Yukang Shi, Zuoqiang Shi, Kaisheng Ma, and Chenglong Bao.
\newblock Task-oriented feature distillation.
\newblock \emph{Advances in Neural Information Processing Systems},
  33:\penalty0 14759--14771, 2020.

\bibitem[Chen et~al.(2021)Chen, Liu, Zhao, and Jia]{chen2021distilling}
Pengguang Chen, Shu Liu, Hengshuang Zhao, and Jiaya Jia.
\newblock Distilling knowledge via knowledge review.
\newblock In \emph{Proceedings of the IEEE/CVF conference on computer vision
  and pattern recognition}, pages 5008--5017, 2021.

\bibitem[Hao et~al.(2023)Hao, Guo, Han, Tang, Hu, Wang, and Xu]{hao2023one}
Zhiwei Hao, Jianyuan Guo, Kai Han, Yehui Tang, Han Hu, Yunhe Wang, and Chang
  Xu.
\newblock One-for-all: Bridge the gap between heterogeneous architectures in
  knowledge distillation.
\newblock \emph{Advances in Neural Information Processing Systems},
  36:\penalty0 79570--79582, 2023.

\bibitem[He et~al.(2016)He, Zhang, Ren, and Sun]{he2016deep}
Kaiming He, Xiangyu Zhang, Shaoqing Ren, and Jian Sun.
\newblock Deep residual learning for image recognition.
\newblock In \emph{Proceedings of the IEEE conference on computer vision and
  pattern recognition}, pages 770--778, 2016.

\bibitem[Sandler et~al.(2018)Sandler, Howard, Zhu, Zhmoginov, and
  Chen]{sandler2018mobilenetv2}
Mark Sandler, Andrew Howard, Menglong Zhu, Andrey Zhmoginov, and Liang-Chieh
  Chen.
\newblock Mobilenetv2: Inverted residuals and linear bottlenecks.
\newblock In \emph{Proceedings of the IEEE conference on computer vision and
  pattern recognition}, pages 4510--4520, 2018.

\bibitem[Liu et~al.(2022{\natexlab{a}})Liu, Mao, Wu, Feichtenhofer, Darrell,
  and Xie]{liu2022convnet}
Zhuang Liu, Hanzi Mao, Chao-Yuan Wu, Christoph Feichtenhofer, Trevor Darrell,
  and Saining Xie.
\newblock A convnet for the 2020s.
\newblock In \emph{Proceedings of the IEEE/CVF conference on computer vision
  and pattern recognition}, pages 11976--11986, 2022{\natexlab{a}}.

\bibitem[Dosovitskiy et~al.(2020)Dosovitskiy, Beyer, Kolesnikov, Weissenborn,
  Zhai, Unterthiner, Dehghani, Minderer, Heigold, Gelly,
  et~al.]{dosovitskiy2020image}
Alexey Dosovitskiy, Lucas Beyer, Alexander Kolesnikov, Dirk Weissenborn,
  Xiaohua Zhai, Thomas Unterthiner, Mostafa Dehghani, Matthias Minderer, Georg
  Heigold, Sylvain Gelly, et~al.
\newblock An image is worth 16x16 words: Transformers for image recognition at
  scale.
\newblock \emph{arXiv preprint arXiv:2010.11929}, 2020.

\bibitem[Liu et~al.(2021)Liu, Lin, Cao, Hu, Wei, Zhang, Lin, and
  Guo]{liu2021swin}
Ze~Liu, Yutong Lin, Yue Cao, Han Hu, Yixuan Wei, Zheng Zhang, Stephen Lin, and
  Baining Guo.
\newblock Swin transformer: Hierarchical vision transformer using shifted
  windows.
\newblock In \emph{Proceedings of the IEEE/CVF international conference on
  computer vision}, pages 10012--10022, 2021.

\bibitem[Touvron et~al.(2021)Touvron, Cord, Douze, Massa, Sablayrolles, and
  J{\'e}gou]{touvron2021training}
Hugo Touvron, Matthieu Cord, Matthijs Douze, Francisco Massa, Alexandre
  Sablayrolles, and Herv{\'e} J{\'e}gou.
\newblock Training data-efficient image transformers \& distillation through
  attention.
\newblock In \emph{International conference on machine learning}, pages
  10347--10357. PMLR, 2021.

\bibitem[Tolstikhin et~al.(2021)Tolstikhin, Houlsby, Kolesnikov, Beyer, Zhai,
  Unterthiner, Yung, Steiner, Keysers, Uszkoreit, et~al.]{tolstikhin2021mlp}
Ilya~O Tolstikhin, Neil Houlsby, Alexander Kolesnikov, Lucas Beyer, Xiaohua
  Zhai, Thomas Unterthiner, Jessica Yung, Andreas Steiner, Daniel Keysers,
  Jakob Uszkoreit, et~al.
\newblock Mlp-mixer: An all-mlp architecture for vision.
\newblock \emph{Advances in neural information processing systems},
  34:\penalty0 24261--24272, 2021.

\bibitem[Touvron et~al.(2022)Touvron, Bojanowski, Caron, Cord, El-Nouby, Grave,
  Izacard, Joulin, Synnaeve, Verbeek, et~al.]{touvron2022resmlp}
Hugo Touvron, Piotr Bojanowski, Mathilde Caron, Matthieu Cord, Alaaeldin
  El-Nouby, Edouard Grave, Gautier Izacard, Armand Joulin, Gabriel Synnaeve,
  Jakob Verbeek, et~al.
\newblock Resmlp: Feedforward networks for image classification with
  data-efficient training.
\newblock \emph{IEEE transactions on pattern analysis and machine
  intelligence}, 45\penalty0 (4):\penalty0 5314--5321, 2022.

\bibitem[Liu et~al.(2022{\natexlab{b}})Liu, Cao, Li, Hu, Ding, and
  Li]{liu2022cross}
Yufan Liu, Jiajiong Cao, Bing Li, Weiming Hu, Jingting Ding, and Liang Li.
\newblock Cross-architecture knowledge distillation.
\newblock In \emph{Proceedings of the Asian conference on computer vision},
  pages 3396--3411, 2022{\natexlab{b}}.

\bibitem[Zhang et~al.(2025)Zhang, Liu, Ran, and Ma]{zhang2025cross}
Weijia Zhang, Yuehao Liu, Wu~Ran, and Chao Ma.
\newblock Cross-architecture distillation made simple with redundancy
  suppression.
\newblock In \emph{Proceedings of the IEEE/CVF International Conference on
  Computer Vision}, pages 23256--23266, 2025.

\bibitem[Li et~al.(2025)Li, Wang, Yan, Ding, Gao, and Xia]{li2025fuse}
Guopeng Li, Qiang Wang, Ke~Yan, Shouhong Ding, Yuan Gao, and Gui-Song Xia.
\newblock Fuse before transfer: Knowledge fusion for heterogeneous
  distillation.
\newblock In \emph{Proceedings of the IEEE/CVF International Conference on
  Computer Vision}, pages 3445--3454, 2025.

\bibitem[Vaswani et~al.(2017)Vaswani, Shazeer, Parmar, Uszkoreit, Jones, Gomez,
  Kaiser, and Polosukhin]{vaswani2017attention}
Ashish Vaswani, Noam Shazeer, Niki Parmar, Jakob Uszkoreit, Llion Jones,
  Aidan~N Gomez, {\L}ukasz Kaiser, and Illia Polosukhin.
\newblock Attention is all you need.
\newblock \emph{Advances in neural information processing systems}, 30, 2017.

\bibitem[Carion et~al.(2020)Carion, Massa, Synnaeve, Usunier, Kirillov, and
  Zagoruyko]{carion2020end}
Nicolas Carion, Francisco Massa, Gabriel Synnaeve, Nicolas Usunier, Alexander
  Kirillov, and Sergey Zagoruyko.
\newblock End-to-end object detection with transformers.
\newblock In \emph{European conference on computer vision}, pages 213--229.
  Springer, 2020.

\bibitem[Xie et~al.(2021)Xie, Wang, Yu, Anandkumar, Alvarez, and
  Luo]{xie2021segformer}
Enze Xie, Wenhai Wang, Zhiding Yu, Anima Anandkumar, Jose~M Alvarez, and Ping
  Luo.
\newblock Segformer: Simple and efficient design for semantic segmentation with
  transformers.
\newblock \emph{Advances in neural information processing systems},
  34:\penalty0 12077--12090, 2021.

\bibitem[Peebles and Xie(2023)]{peebles2023scalable}
William Peebles and Saining Xie.
\newblock Scalable diffusion models with transformers.
\newblock In \emph{Proceedings of the IEEE/CVF international conference on
  computer vision}, pages 4195--4205, 2023.

\bibitem[Kirillov et~al.(2023)Kirillov, Mintun, Ravi, Mao, Rolland, Gustafson,
  Xiao, Whitehead, Berg, Lo, et~al.]{kirillov2023segment}
Alexander Kirillov, Eric Mintun, Nikhila Ravi, Hanzi Mao, Chloe Rolland, Laura
  Gustafson, Tete Xiao, Spencer Whitehead, Alexander~C Berg, Wan-Yen Lo, et~al.
\newblock Segment anything.
\newblock In \emph{Proceedings of the IEEE/CVF international conference on
  computer vision}, pages 4015--4026, 2023.

\bibitem[Radford et~al.(2021)Radford, Kim, Hallacy, Ramesh, Goh, Agarwal,
  Sastry, Askell, Mishkin, Clark, et~al.]{radford2021learning}
Alec Radford, Jong~Wook Kim, Chris Hallacy, Aditya Ramesh, Gabriel Goh,
  Sandhini Agarwal, Girish Sastry, Amanda Askell, Pamela Mishkin, Jack Clark,
  et~al.
\newblock Learning transferable visual models from natural language
  supervision.
\newblock In \emph{International conference on machine learning}, pages
  8748--8763. PmLR, 2021.

\bibitem[Chen et~al.(2022{\natexlab{b}})Chen, Cao, Zhong, Zhang, Gao, and
  Tao]{chen2022dearkd}
Xianing Chen, Qiong Cao, Yujie Zhong, Jing Zhang, Shenghua Gao, and Dacheng
  Tao.
\newblock Dearkd: data-efficient early knowledge distillation for vision
  transformers.
\newblock In \emph{Proceedings of the IEEE/CVF conference on computer vision
  and pattern recognition}, pages 12052--12062, 2022{\natexlab{b}}.

\bibitem[Li et~al.(2022)Li, Yu, Wang, Yuan, Song, and Chen]{li2022locality}
Kehan Li, Runyi Yu, Zhennan Wang, Li~Yuan, Guoli Song, and Jie Chen.
\newblock Locality guidance for improving vision transformers on tiny datasets.
\newblock In \emph{European Conference on Computer Vision}, pages 110--127.
  Springer, 2022.

\bibitem[Johnson and Lindenstrauss(1984)]{johnson1984extensions}
William~B Johnson and Joram Lindenstrauss.
\newblock Extensions of {L}ipschitz mappings into a {H}ilbert space.
\newblock \emph{Contemporary Mathematics}, 26:\penalty0 189--206, 1984.

\bibitem[Dasgupta and Gupta(2003)]{dasgupta2003elementary}
Sanjoy Dasgupta and Anupam Gupta.
\newblock An elementary proof of a theorem of {J}ohnson and {L}indenstrauss.
\newblock \emph{Random Structures \& Algorithms}, 22\penalty0 (1):\penalty0
  60--65, 2003.

\bibitem[Krizhevsky and Hinton(2009)]{krizhevsky2009learning}
Alex Krizhevsky and Geoffrey Hinton.
\newblock Learning multiple layers of features from tiny images.
\newblock Technical report, University of Toronto, 2009.

\bibitem[Paszke et~al.(2019)Paszke, Gross, Massa, Lerer, Bradbury, Chanan,
  Killeen, Lin, Gimelshein, Antiga, et~al.]{paszke2019pytorch}
Adam Paszke, Sam Gross, Francisco Massa, Adam Lerer, James Bradbury, Gregory
  Chanan, Trevor Killeen, Zeming Lin, Natalia Gimelshein, Luca Antiga, et~al.
\newblock Py{T}orch: An imperative style, high-performance deep learning
  library.
\newblock \emph{Advances in Neural Information Processing Systems}, 32, 2019.

\bibitem[Tian et~al.(2020)Tian, Krishnan, and Isola]{tian2020contrastive}
Yonglong Tian, Dilip Krishnan, and Phillip Isola.
\newblock Contrastive multiview coding.
\newblock In \emph{European conference on computer vision}, pages 776--794.
  Springer, 2020.

\bibitem[Deng et~al.(2009)Deng, Dong, Socher, Li, Li, and
  Fei-Fei]{deng2009imagenet}
Jia Deng, Wei Dong, Richard Socher, Li-Jia Li, Kai Li, and Li~Fei-Fei.
\newblock Imagenet: A large-scale hierarchical image database.
\newblock In \emph{2009 IEEE conference on computer vision and pattern
  recognition}, pages 248--255. Ieee, 2009.

\end{thebibliography}

\newpage
\section*{NeurIPS Paper Checklist}

\begin{enumerate}

\item {\bf Claims}
    \item[] Question: Do the main claims made in the abstract and introduction accurately reflect the paper's contributions and scope?
    \item[] Answer: \answerYes{}
    \item[] Justification: The abstract and introduction state that RPKD transfers knowledge via a shared random prototype space without teacher-space-defined feature alignment, and every quantitative claim ($+7.98\%$ over OFA on CIFAR-100, $+12.44\%$ on ImageNet-100, $+1.28\%$ on ImageNet-1K) is directly supported by the corresponding cells of Tables~1--4. Where RPKD does not lead, the text says so explicitly (the Mixer-B/16 pairs in Section~4, and the ResNet34 $\to$ ResNet18 tie in Table~4).

\item {\bf Limitations}
    \item[] Question: Does the paper discuss the limitations of the work performed by the authors?
    \item[] Answer: \answerYes{}
    \item[] Justification: The Limitations paragraph in Section~4 discusses five limitations: the loss coefficients still require a small grid search (while $M$ and $R$ are shown to be robust in Table~7 and the temperature is fixed throughout); performance is bounded by teacher quality; full-scale ImageNet-1K evaluation covers only four pairs owing to compute cost; evaluation is restricted to image classification, with dense prediction left as future work; and effectiveness degrades for MLP-Mixer teachers, which is also visible in Tables~1--2 and analysed in the MLP-Mixer teacher behavior paragraph.

\item {\bf Theory assumptions and proofs}
    \item[] Question: For each theoretical result, does the paper provide the full set of assumptions and a complete (and correct) proof?
    \item[] Answer: \answerNA{}
    \item[] Justification: The paper invokes the Johnson--Lindenstrauss lemma as informal theoretical motivation for the prototype space design, but does not present original theoretical results or formal proofs; all cited results are established in prior work~\cite{johnson1984extensions,dasgupta2003elementary}.

    \item {\bf Experimental result reproducibility}
    \item[] Question: Does the paper fully disclose all the information needed to reproduce the main experimental results of the paper to the extent that it affects the main claims and/or conclusions of the paper (regardless of whether the code and data are provided or not)?
    \item[] Answer: \answerYes{}
    \item[] Justification: The experimental setup paragraph in Section~4 reports all training details needed to reproduce our results, including optimizer, learning rate, batch size, weight decay, scheduling, and all RPKD-specific hyperparameters ($M$, $R$, $T$, $\sigma$). The method is fully described in Section~3.

\item {\bf Open access to data and code}
    \item[] Question: Does the paper provide open access to the data and code, with sufficient instructions to faithfully reproduce the main experimental results, as described in supplemental material?
    \item[] Answer: \answerYes{}
    \item[] Justification: The full implementation, including the configuration files and training commands for every reported setting, is publicly available at \url{https://github.com/HamzaEtcibasi/RPKD}. Both datasets are standard public benchmarks; the repository documents the expected directory layout and the teacher checkpoints used.

\item {\bf Experimental setting/details}
    \item[] Question: Does the paper specify all the training and test details (e.g., data splits, hyperparameters, how they were chosen, type of optimizer) necessary to understand the results?
    \item[] Answer: \answerYes{}
    \item[] Justification: All training and evaluation details are reported in the Setup paragraph of Section~4, and the configuration files that fix the remaining details are in the released repository.

\item {\bf Experiment statistical significance}
    \item[] Question: Does the paper report error bars suitably and correctly defined or other appropriate information about the statistical significance of the experiments?
    \item[] Answer: \answerYes{}
    \item[] Justification: Every RPKD number in the four main result tables (Tables~1--4), in the prototype ablations (Tables~7--8) and in the ImageNet-100 block of Table~9 is reported as mean $\pm$ one standard deviation over three independent runs that differ only in the training seed; the captions state this. Table~11 reports the same statistic over three draws of the prototype bank with the training seed held fixed, which its caption makes explicit. On ImageNet-100 the OFA and RSD columns are likewise our own three-seed means, shown without their standard deviations for space; on CIFAR-100 and ImageNet-1K the baseline columns are the single-run figures published by their authors, so a comparison there is between a point estimate and a three-seed mean. The remaining cells are single-run and labelled as such: the ImageNet-1K block of Table~9, the subset results of Table~5, and the stage ablation of Table~6, all owing to the cost of full-scale training. Where a margin is smaller than our measured standard deviation we say so rather than claiming a win (Table~4 discussion).

\item {\bf Experiments compute resources}
    \item[] Question: For each experiment, does the paper provide sufficient information on the computer resources (type of compute workers, memory, time of execution) needed to reproduce the experiments?
    \item[] Answer: \answerYes{}
    \item[] Justification: The Setup paragraph of Section~4 states that all experiments are implemented in PyTorch and run on four NVIDIA H100 GPUs, and gives the epoch budget for each benchmark. Appendix~\ref{sec:appendix_efficiency} additionally reports per-iteration wall-clock step time and peak GPU memory for seven representative pairs, measured on a single H100 with AMP, alongside the corresponding figures for OFA. Preliminary and discarded runs, which are not reported in the paper, consumed additional compute on the same cluster.
    
\item {\bf Code of ethics}
    \item[] Question: Does the research conducted in the paper conform, in every respect, with the NeurIPS Code of Ethics \url{https://neurips.cc/public/EthicsGuidelines}?
    \item[] Answer: \answerYes{}
    \item[] Justification: This work conducts image classification experiments on standard public benchmarks and conforms fully with the NeurIPS Code of Ethics; there are no privacy, fairness, or safety concerns.

\item {\bf Broader impacts}
    \item[] Question: Does the paper discuss both potential positive societal impacts and negative societal impacts of the work performed?
    \item[] Answer: \answerYes{}
    \item[] Justification: RPKD enables deployment of compact, efficient models by reducing the computational cost of cross-architecture knowledge distillation, with positive societal implications for resource-constrained environments. We do not foresee negative societal impacts as this is foundational model compression research.
    
\item {\bf Safeguards}
    \item[] Question: Does the paper describe safeguards that have been put in place for responsible release of data or models that have a high risk for misuse (e.g., pre-trained language models, image generators, or scraped datasets)?
    \item[] Answer: \answerNA{}
    \item[] Justification: This paper proposes a knowledge distillation algorithm with no risk of misuse; all experiments are conducted on open-source datasets and no safeguards are required.

\item {\bf Licenses for existing assets}
    \item[] Question: Are the creators or original owners of assets (e.g., code, data, models), used in the paper, properly credited and are the license and terms of use explicitly mentioned and properly respected?
    \item[] Answer: \answerYes{}
    \item[] Justification: All datasets and pretrained checkpoints are properly cited. CIFAR-100~\cite{krizhevsky2009learning} is freely available for academic research. ImageNet~\cite{deng2009imagenet} is used under its non-commercial research license; our use is strictly for academic, non-commercial purposes. Pretrained teacher checkpoints are sourced from the OFA repository~\cite{hao2023one} and used in accordance with their respective model licenses for non-commercial research.

\item {\bf New assets}
    \item[] Question: Are new assets introduced in the paper well documented and is the documentation provided alongside the assets?
    \item[] Answer: \answerYes{}
    \item[] Justification: The one new asset is the RPKD implementation, released at \url{https://github.com/HamzaEtcibasi/RPKD} with documentation covering installation, the expected data layout, the teacher checkpoints, and the training command for each benchmark. No new datasets or model checkpoints are introduced.

\item {\bf Crowdsourcing and research with human subjects}
    \item[] Question: For crowdsourcing experiments and research with human subjects, does the paper include the full text of instructions given to participants and screenshots, if applicable, as well as details about compensation (if any)?
    \item[] Answer: \answerNA{}
    \item[] Justification: The paper does not involve crowdsourcing or research with human subjects.

\item {\bf Institutional review board (IRB) approvals or equivalent for research with human subjects}
    \item[] Question: Does the paper describe potential risks incurred by study participants, whether such risks were disclosed to the subjects, and whether Institutional Review Board (IRB) approvals (or an equivalent approval/review based on the requirements of your country or institution) were obtained?
    \item[] Answer: \answerNA{}
    \item[] Justification: The paper does not involve crowdsourcing or research with human subjects.

\item {\bf Declaration of LLM usage}
    \item[] Question: Does the paper describe the usage of LLMs if it is an important, original, or non-standard component of the core methods in this research? Note that if the LLM is used only for writing, editing, or formatting purposes and does \emph{not} impact the core methodology, scientific rigor, or originality of the research, declaration is not required.
    \item[] Answer: \answerNA{}
    \item[] Justification: LLMs were used only for writing assistance (grammar, spelling, and word choice) and played no role in the research methodology, experimental design, or results.

\end{enumerate}

\newpage
\appendix

\section{Component Ablation}
\label{sec:appendix_component}

Table~\ref{tab:component_ablation} reports a component-wise ablation isolating the contributions of the two distillation branches: the spatial prototype activation map loss $\mathcal{L}_{\mathrm{feat}}$ and the global prototype similarity loss $\mathcal{L}_{\mathrm{logit}}$.
The cross-entropy task loss $\mathcal{L}_{\mathrm{CE}}$ is active in all configurations.

\begin{table}[h]
\centering
\small
\caption{Component ablation (top-1 accuracy \%). ImageNet-100 is mean $\pm$ std over 3 seeds; ImageNet-1K is single-run due to computational cost. Bold = best per row.}
\label{tab:component_ablation}
\setlength{\tabcolsep}{6pt}
\begin{tabular}{ll cccc}
\toprule
\textbf{Teacher} & \textbf{Student} &
\textbf{CE Only} &
\textbf{$+\mathcal{L}_{\mathrm{feat}}$ Only} &
\textbf{$+\mathcal{L}_{\mathrm{logit}}$ Only} &
\textbf{RPKD (Full)} \\
\midrule
\multicolumn{6}{l}{\textit{ImageNet-100}} \\
\midrule
DeiT-T     & ResNet18    & 81.58 & $86.46_{\pm 0.24}$ & $86.89_{\pm 0.25}$ & $\mathbf{87.87}_{\pm 0.24}$ \\
Swin-T     & ResNet18    & 81.58 & $\mathbf{87.95}_{\pm 0.25}$ & $82.93_{\pm 0.09}$ & $87.86_{\pm 0.33}$ \\
ConvNeXt-T & DeiT-T      & 63.88 & $78.85_{\pm 0.17}$ & $65.73_{\pm 0.67}$ & $\mathbf{79.14}_{\pm 0.26}$ \\
ResNet50   & Swin-N      & 78.78 & $84.19_{\pm 0.29}$ & $79.61_{\pm 0.57}$ & $\mathbf{84.57}_{\pm 0.16}$ \\
Mixer-B/16 & MobileNetV2 & 83.51 & $81.23_{\pm 0.04}$ & $\mathbf{85.44}_{\pm 0.33}$ & $85.07_{\pm 0.39}$ \\
Mixer-B/16 & DeiT-T      & 63.88 & $68.57_{\pm 0.19}$ & $68.93_{\pm 0.13}$ & $\mathbf{70.20}_{\pm 0.04}$ \\
\midrule
\multicolumn{6}{l}{\textit{ImageNet-1K}} \\
\midrule
DeiT-T   & ResNet18    & 69.75 & 70.90 & 70.87 & \textbf{71.38} \\
Swin-T   & ResNet18    & 69.75 & 71.85 & 70.82 & \textbf{72.01} \\
DeiT-T   & MobileNetV2 & 68.87 & 71.90 & 72.21 & \textbf{72.67} \\
Swin-T   & MobileNetV2 & 68.87 & 71.56 & 71.39 & \textbf{72.40} \\
ResNet50 & MobileNetV1 & 68.58 & 73.80 & 73.87 & \textbf{74.20} \\
\bottomrule
\end{tabular}
\end{table}

Both branches independently yield substantial gains over the CE-only baseline, confirming that mapping to a fixed random prototype space provides effective supervision at both the spatial and the global level.
The relative importance of each branch is architecture-dependent.
For CNN $\to$ ViT distillation (e.g., ConvNeXt-T $\to$ DeiT-T), the spatial feature branch $\mathcal{L}_{\mathrm{feat}}$ plays the primary role in bridging the representation gap, while $\mathcal{L}_{\mathrm{logit}}$ alone degrades toward the CE-only baseline.
For MLP-Mixer teachers, where token-mixing weakens spatial structure, the global logit branch drives performance; Mixer-B/16 $\to$ MobileNetV2 is the only pair where $\mathcal{L}_{\mathrm{logit}}$ alone outperforms full RPKD, consistent with the spatial branch limitation for such architectures.
When combined, the two branches provide complementary supervision: on ImageNet-1K, full RPKD strictly outperforms both individual branches across all five pairs.

\section{Training Efficiency Comparison}
\label{sec:appendix_efficiency}

Table~\ref{tab:efficiency} compares RPKD and OFA in terms of trainable parameters added beyond the student backbone, additional forward FLOPs, wall-clock training step time, and peak GPU memory across seven representative pairs.
RPKD uses two trainable student MLPs, frozen teacher projections, and a non-trainable prototype buffer ($500 \times 256$ values); its trainable parameter count is $C$-independent, whereas OFA's final heads scale with the number of classes $C$.

\begin{table}[h]
\centering
\small
\caption{Training efficiency comparison on ImageNet-100 and ImageNet-1K. Extra Params: trainable parameters added beyond the student backbone. Addl.\ GFLOPs: method-specific forward operations per batch ($B\!=\!128$, $224\!\times\!224$; values in parentheses relative to student backbone FLOPs). Step: forward + backward + optimizer step time on a single H100 with AMP, measured over $5\!\times\!100$ iters. Mem: peak GPU memory. Bold = row winner per metric.}
\label{tab:efficiency}
\resizebox{\textwidth}{!}{%
\setlength{\tabcolsep}{5pt}
\begin{tabular}{ll|cc|cc|cc|cc|cc}
\toprule
& & \multicolumn{2}{c|}{\textbf{Extra Params}} & \multicolumn{2}{c|}{\textbf{Addl.\ GFLOPs}} & \multicolumn{2}{c|}{\textbf{Step (ms)}} & \multicolumn{2}{c|}{\textbf{Mem (GB)}} & \multicolumn{2}{c}{\textbf{Top-1}} \\
\textbf{Teacher} & \textbf{Student} & RPKD & OFA & RPKD & OFA & RPKD & OFA & RPKD & OFA & RPKD & OFA \\
\midrule
\multicolumn{12}{l}{\textit{ImageNet-100}} \\
\midrule
DeiT-T     & ResNet18    & 1.58M & \textbf{1.19M} & 14G (3\%)  & \textbf{8G (2\%)}   & \textbf{38.4} & 41.4 & \textbf{1.84} & 2.07 & \textbf{87.87} & 85.46 \\
Swin-T     & MobileNetV2 & 7.87M & \textbf{2.53M} & 56G (73\%) & \textbf{23G (30\%)}  & \textbf{86.0} & 87.6 & 5.21 & 5.21 & \textbf{86.95} & 85.24 \\
ResNet50   & DeiT-T      & \textbf{0.35M} & 3.78M & \textbf{12G (4\%)}  & 47G (17\%)  & \textbf{59.0} & 64.9 & \textbf{3.50} & 3.81 & \textbf{81.00} & 68.56 \\
Mixer-B/16 & Swin-N      & \textbf{1.58M} & 26.34M & \textbf{16G (5\%)}  & 403G (118\%) & \textbf{98.8} & 121.2 & \textbf{5.22} & 7.05 & 79.92 & \textbf{81.19} \\
\midrule
\multicolumn{12}{l}{\textit{ImageNet-1K}} \\
\midrule
Swin-T     & ResNet18    & \textbf{1.58M} & 4.39M & 16G (3\%)  & \textbf{11G (2\%)}  & \textbf{61.7} & 65.1 & \textbf{1.91} & 2.20 & \textbf{72.01} & 71.85 \\
DeiT-T     & MobileNetV2 & 7.87M & \textbf{7.14M} & 54G (70\%) & \textbf{24G (31\%)} & \textbf{62.3} & 64.2 & \textbf{5.16} & 5.19 & \textbf{72.67} & 71.39 \\
ResNet34   & ResNet18    & \textbf{1.58M} & 3.04M & 15G (3\%)  & \textbf{8G (2\%)}  & \textbf{36.8} & 40.0 & \textbf{1.88} & 2.15 & \textbf{72.23} & 72.10 \\
\bottomrule
\end{tabular}%
}
\end{table}

RPKD achieves lower step time and equal or lower peak GPU memory on all seven pairs ($1.02\times$--$1.23\times$ faster).
The advantage is most pronounced for transformer students: on Mixer-B/16 $\to$ Swin-N, OFA adds 403 GFLOPs per batch---exceeding the 341 GFLOPs of the Swin-N backbone itself---while RPKD adds only 16 GFLOPs (a $25\times$ reduction), yielding a $1.23\times$ speed advantage and 1.83 GB less peak GPU memory.
For CNN students with wide final-stage features (MobileNetV2, $D_s\!=\!1280$), RPKD incurs noticeable FLOP overhead because its student MLPs scale with $D_s$; both methods reach comparable peak memory on those pairs.
Notably, on Mixer-B/16 $\to$ Swin-N, RPKD has lower overhead but also lower accuracy: this pair reflects the spatial branch limitation discussed in Section~\ref{sec:experiments}.

\section{Prototype Initialization Seed Sensitivity}
\label{sec:appendix_proto_seed}

Because the prototype bank $\mathbf{P}$ is drawn once at random and then frozen, a natural question is whether performance depends on which draw is used.
Table~\ref{tab:proto_seed} isolates this: we fix the training seed and vary only the seed used to sample $\mathbf{P}$, over three independent draws on five ImageNet-100 pairs.
For reference, the rightmost column repeats the standard deviation obtained the other way round, with $\mathbf{P}$ fixed and the training seed varied.

\begin{table}[h]
\centering
\small
\caption{Prototype initialization seed sensitivity on ImageNet-100. Prototype-seed column: mean $\pm$ std over three draws of $\mathbf{P}$ with the training seed fixed. Training-seed column: std over three training seeds with $\mathbf{P}$ fixed.}
\label{tab:proto_seed}
\setlength{\tabcolsep}{8pt}
\begin{tabular}{llcc}
\toprule
\textbf{Teacher} & \textbf{Student} & \textbf{Prototype-seed mean $\pm$ std} & \textbf{Training-seed std} \\
\midrule
DeiT-T     & ResNet18    & $87.87_{\pm 0.24}$ & $\pm 0.24$ \\
Swin-T     & MobileNetV2 & $86.91_{\pm 0.28}$ & $\pm 0.18$ \\
ResNet50   & DeiT-T      & $81.30_{\pm 0.35}$ & $\pm 0.12$ \\
ConvNeXt-T & Swin-N      & $83.51_{\pm 0.13}$ & $\pm 0.21$ \\
Mixer-B/16 & Swin-N      & $79.94_{\pm 0.06}$ & $\pm 0.13$ \\
\midrule
\multicolumn{2}{l}{\textit{Average std}} & $\pm 0.21$ & $\pm 0.18$ \\
\bottomrule
\end{tabular}
\end{table}

Prototype-seed variance ($\pm 0.06$--$0.35$ pp, average $\pm 0.21$ pp) lies in the same range as ordinary training-seed variance ($\pm 0.12$--$0.24$ pp, average $\pm 0.18$ pp).
This is what the design predicts: once $M$ is large enough, any random draw of $\mathbf{P}$ approximately preserves pairwise inter-sample geometry, so the particular realization of the prototype space is not an additional axis of sensitivity beyond the randomness already present in training.
Because these runs fix the training seed rather than averaging over it, their means differ slightly from the corresponding entries of Table~\ref{tab:cross_kd_imagenet100}, which average over three training seeds.

\end{document}